\documentclass[11pt]{article}

\usepackage[margin=1in]{geometry}

\usepackage{microtype}
\usepackage{graphicx}
\usepackage{subfigure}
\usepackage{booktabs}
\usepackage{multirow}
\usepackage{enumitem}

\usepackage[section]{placeins}
\usepackage{float}
\makeatletter
\renewcommand{\fps@figure}{htbp}
\renewcommand{\fps@table}{htbp}
\makeatother

\usepackage[round,authoryear,sort&compress]{natbib}
\usepackage{hyperref}
\hypersetup{
    colorlinks=true,
    linkcolor=blue,
    citecolor=blue,
    urlcolor=blue
}

\usepackage{amsmath}
\usepackage{amssymb}
\usepackage{mathtools}
\usepackage{amsthm}

\usepackage[capitalize,noabbrev]{cleveref}

\usepackage{authblk}

\usepackage{tcolorbox}
\tcbuselibrary{skins}

\theoremstyle{plain}

\theoremstyle{definition}

\theoremstyle{remark}

\usepackage[textsize=tiny]{todonotes}

\usepackage{dblfloatfix}
\usepackage{caption}

\makeatletter
\renewcommand{\paragraph}[1]{\noindent\textbf{#1}}
\makeatother

\begin{document}

\title{\textbf{Mix, MinHash, and Match}\\Cross-Source Agreement for Multilingual Pretraining Datasets}

\author{Sultan Alrashed\textsuperscript{\dag}}
\author{Francesco Orabona}
\affil{\texttt{sultan.alrashed@kaust.edu.sa}, \texttt{francesco@orabona.com}\\King Abdullah University of Science and Technology (KAUST)\\Thuwal, 23955-6900, Kingdom of Saudi Arabia}

\date{}

\maketitle

\renewcommand{\thefootnote}{\fnsymbol{footnote}}
\footnotetext[2]{Corresponding author}
\renewcommand{\thefootnote}{\arabic{footnote}}

\begin{tcolorbox}[
    colback=gray!10,
    colframe=gray!50,
    arc=4mm,
    boxrule=0.5pt,
    left=10pt,
    right=10pt,
    top=10pt,
    bottom=10pt
]
\noindent\textbf{Abstract.}
Multilingual data from the web is essential for LLM pretraining. Yet, scraping it is expensive, and research groups repeatedly crawl the same content. For example, we found that over 40\% of tokens across major Arabic web corpora are duplicated between sources. In this work, we propose to use this wasteful redundancy as a quality signal to create high-quality pretraining datasets. Our key insight is that cross-source agreement functions as a free, model-free quality filter: content retained by multiple independent pipelines is more likely to represent high-quality text. Crucially, this signal requires no additional computation beyond standard deduplication, which is already performed at scale when pretraining language models. So, we propose MixMinMatch, a method that combines multiple existing web corpora, performs cross-dataset MinHash deduplication, and identifies documents independently recovered by multiple sources. We apply MixMinMatch to Arabic, Turkish, and Hindi, producing corpora that match or exceed the quality of the best single-source baselines, while providing up to 4$\times$ more unique tokens. On Arabic, our matched subset achieves a 4.5\% relative improvement over ArabicWeb24, while on Turkish, we improve over FineWeb-2 by 5.5\%. We release the datasets at: \url{https://huggingface.co/collections/AdaMLLab/mixminmatch}
\end{tcolorbox}

\vspace{1em}

\begin{figure}[H]
\centering
\includegraphics[width=0.8\textwidth]{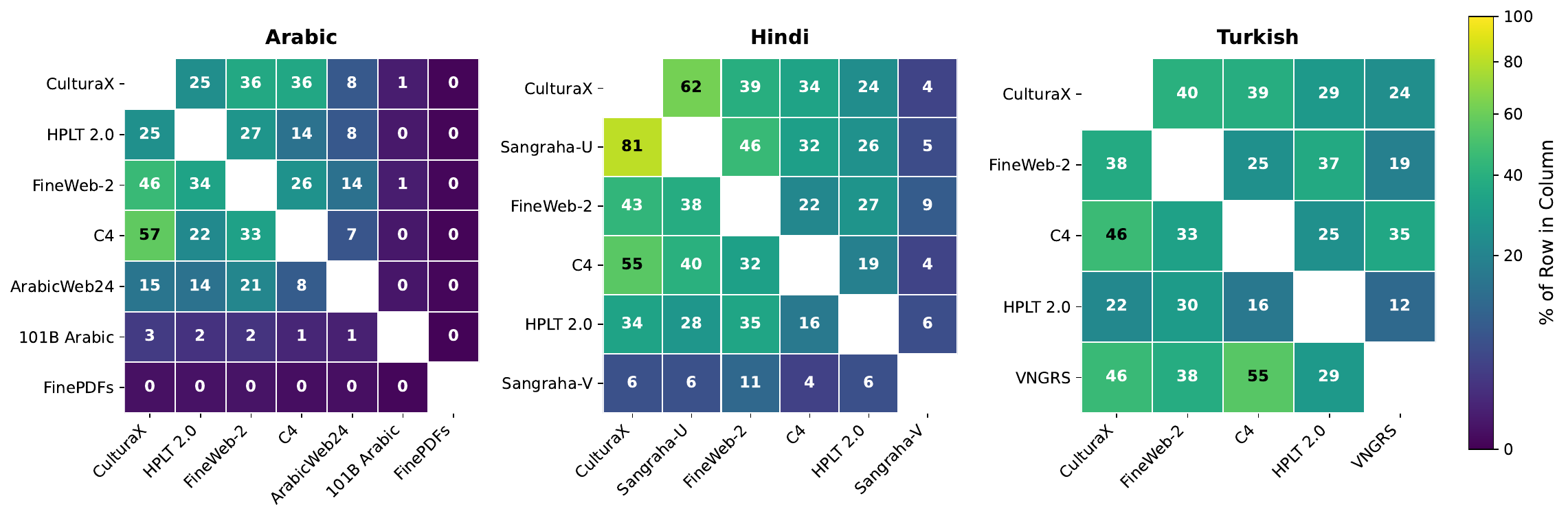}
\caption{Pairwise token overlap between source datasets after quality filtering. Values represent the percent of tokens in row datasets present in column datasets.}
\label{fig:pairwise}
\end{figure}

\section{Introduction}

Building pretraining corpora through web scraping requires substantial computational resources, storage infrastructure, and engineering effort. For languages beyond English, multiple research groups have independently undertaken this expensive process, producing overlapping datasets that duplicate both the crawling work and the crawled content itself. At first glance, this redundancy represents pure inefficiency: resources spent re-scraping the same web pages could instead fund better filtering or additional data collection.

We argue for a different perspective: \emph{this redundancy is itself a signal of quality}. When multiple independent research groups, each with different crawling schedules, filtering pipelines, and quality heuristics, all retain the same document, this cross-source agreement provides evidence that the content is valuable. Conversely, content that appears in only one corpus may have been filtered out by others for good reason, or may represent ephemeral web content captured by a single crawl snapshot.

This insight connects to a foundational principle in machine learning and statistics: independent agreement reduces uncertainty. In crowdsourcing, inter-annotator agreement indicates label reliability~\citep{dawid1979maximum}. In ensemble methods, aggregating independent predictions reduces variance and improves accuracy~\citep{breiman1996bagging}. We apply this same principle to pretraining data: cross-source agreement serves as an ensemble filter, where each dataset's processing pipeline casts an independent ``vote'' on content quality.

Five major multilingual efforts, C4~\citep{raffel2020exploring}, CulturaX~\citep{nguyen2024culturax}, HPLT~2.0~\citep{burchell2025hpltv2}, FinePDFs~\citep{huggingfacefw2025finepdfs}, and FineWeb-2~\citep{penedo2025fineweb2}, now cover dozens of languages, ensuring that most languages have at least five independent sources available for cross-source methods. Arabic goes further with at least seven public corpora: the five above plus ArabicWeb24~\citep{farhat2024arabicweb24} and the ClusterLab 101B Arabic Words dataset~\citep{aloui2024arabic101b}. As Figure~\ref{fig:pairwise} shows, these corpora exhibit substantial pairwise overlap; for instance, C4 and CulturaX share over 9 billion tokens of near-duplicate content. Each overlap represents content that two independent teams, with different goals and methods, both deemed worth retaining.

We propose MixMinMatch, a three-stage method for exploiting this structure:
\begin{enumerate}
    \item \textbf{Mix:} Aggregate multiple existing corpora while preserving the source provenance.
    \item \textbf{MinHash:} Perform cross-dataset near-duplicate detection using locality-sensitive hashing, producing deduplicated clusters.
    \item \textbf{Match:} Retain documents whose near-duplicate clusters span at least two independent sources, using cross-source agreement as a quality signal.
\end{enumerate}

A critical advantage of this approach is its computational efficiency. MinHash deduplication is already standard practice when combining multiple corpora for pretraining, without it, models would train on near-duplicate content, wasting compute and potentially degrading performance~\citep{lee2022deduplicating}. Hence, MixMinMatch requires \emph{no additional computation} beyond this standard deduplication: cross-source matching simply examines the source labels within clusters that MinHash already produces. In contrast, model-based quality filtering~\citep{penedo2024fineweb} requires $O(n)$ inference passes through a quality classifier or language model. Instead, our method extracts a quality signal essentially for free.

Our contributions are:
\begin{itemize}
    \item A systematic analysis of redundancy across multilingual web corpora, quantifying cross-source overlap for four languages.
    \item MixMinMatch, a method that extracts a free quality signal from standard deduplication infrastructure by treating cross-source agreement as ensemble filtering.
    \item Four multilingual pretraining corpora (AraMix, TurMix, HinMix) with per-document source counts, enabling practitioners to tune parameters.
    \item Experimental evidence that cross-source matching identifies high-quality content, achieving competitive or superior performance to single-source baselines when pretraining.
\end{itemize}

\section{Related Work}

\noindent\textbf{Corpus Aggregation and Deduplication.} Prior work has combined and deduplicated existing corpora, though typically within narrower scope. CulturaX~\citep{nguyen2024culturax} merged mC4~\citep{raffel2020exploring} and OSCAR~\citep{ortiz2019oscar} with additional cleaning but did not leverage cross-source overlap as a quality signal. SlimPajama~\citep{soboleva2023slimpajama} applied MinHash deduplication to RedPajama~\citep{together2023redpajama}, removing 49\% of content and demonstrating substantial within-corpus duplication. \citet{lee2022deduplicating} showed that deduplication improves training efficiency and model quality, establishing it as standard practice. The FineWeb effort~\citep{penedo2024fineweb} analyzed overlap between their corpus and RefinedWeb~\citep{penedo2023refinedweb}, finding significant shared content from Common Crawl. \citet{tirumala2023d4} demonstrated that deduplication combined with diversity-based selection improves training efficiency by 20\%. We extend this line of work by showing that cross-source clusters produced during deduplication encode quality information that prior methods discard.

\noindent\textbf{Language-Specific Quality Filtering.} Quality filters developed for English often fail on morphologically rich or non-Latin-script languages: C4 rejects lines lacking English terminal punctuation, and the Gopher filters~\citep{rae2021gopher} require English stop words. CCNet~\citep{wenzek2020ccnet} addressed this using perplexity filtering against Wikipedia language models, while FineWeb-2~\citep{penedo2025fineweb2} adapted thresholds but retained structurally English-centric heuristics. Here, we apply language-specific filters where necessary, but emphasize that cross-source matching provides a language-agnostic quality signal requiring no such engineering.

\noindent\textbf{Data Quality through Ensemble Agreement.} The principle that independent agreement signals quality has deep roots in statistics and machine learning. In crowdsourcing, inter-annotator agreement metrics like Fleiss' kappa~\citep{fleiss1971measuring} and Krippendorff's alpha~\citep{krippendorff2011computing} quantify label reliability by measuring consensus among independent annotators. The Dawid-Skene model formalizes this intuition, inferring true labels by aggregating noisy annotations. In ensemble learning, bagging and boosting~\citep{freund1997decision} improve predictions by combining independent models, exploiting the fact that independent errors tend to cancel. Our work applies this principle to pretraining data curation: each dataset's filtering pipeline acts as an independent annotator, and cross-source agreement identifies content that multiple ``annotators'' endorse.

\section{Cross-Source Overlap}

Existing multilingual corpora contain substantial overlapping content, representing redundant effort in independent web scraping attempts. First of all, we quantify this redundancy to motivate our approach. Throughout this work, all token counts use \texttt{meta-llama/Llama-3.2-3B}'s tokenizer~\citep{grattafiori2024llama3} to ensure consistency.

Figure~\ref{fig:pairwise} shows pairwise token overlap between major Arabic web corpora after quality filtering. The overlaps are substantial: C4 and CulturaX share over 9 billion tokens of near-duplicate content, while FineWeb-2 and HPLT~2.0 share over 5 billion. Each overlap represents content that two independent teams, with different goals and methods, both deemed worth retaining.

\begin{table}[h]
\centering
\caption{Aggregate token counts by cross-source overlap degree in the matched subset. Overlap degree indicates appearance in unique source datasets.}
\label{tab:overlap_by_degree}
\small
\begin{tabular}{@{}l r r r r r@{}}
\toprule
\textbf{Language} & \textbf{2-way} & \textbf{3-way} & \textbf{4-way} & \textbf{5-way} & \textbf{6-way} \\
\midrule
Arabic  & 28.9B & 16.4B & 7.1B & 1.3B & 32.4M \\
Turkish & 26.0B & 14.3B & 9.5B & 5.7B & --- \\
Hindi   & 10.9B &  8.0B & 5.4B & 1.4B & 0.2B \\
\bottomrule
\end{tabular}
\end{table}

Table~\ref{tab:overlap_by_degree} extends this analysis to higher-order overlaps. For Arabic, 28.9B tokens appear in exactly two sources, with substantial content extending to three, four, and even five independent sources. This pattern holds across all languages we study: the content recovered by multiple pipelines collectively represents tens of billions of tokens per language. We identify the largest pairwise overlaps alongside more detailed statistics in Appendix~\ref{sec:pairwise_appendix}.

In the following, we will show how to use this wasted overlap as a quality signal waiting to be exploited.

\section{The MixMinMatch Algorithm}

The substantial cross-source overlap suggests that independent filtering pipelines frequently agree on content quality. We now describe MixMinMatch, a method that exploits this structure by combining existing corpora, performing joint deduplication, and using cross-source agreement as a quality signal. 

\begin{table*}[t]
\centering
\caption{Token counts at each processing stage across languages. Reduction ratios show the percentage of tokens retained relative to the previous stage. B=Billions, M=Millions.}
\label{tab:processing_stages}
\small
\begin{tabular}{@{}l r r r r r r@{}}
\toprule
\textbf{Language} & \multicolumn{2}{c}{\textbf{Quality Filtered}} & \multicolumn{2}{c}{\textbf{MinHash Deduped}} & \multicolumn{2}{c}{\textbf{Matched}} \\
\cmidrule(lr){2-3} \cmidrule(lr){4-5} \cmidrule(lr){6-7}
 & Tokens & Docs & Tokens & Docs & Tokens & Docs \\
\midrule
Arabic  & 300.9B & 283.4M & 177.8B & 178.9M & 54.1B & 47.9M \\
Turkish & 307.2B & 394.0M & 167.6B & 219.1M & 56.0B & 67.6M \\
Hindi   & 130.3B &  99.6M &  76.2B &  59.6M & 27.1B & 19.8M \\
\bottomrule
\end{tabular}
\end{table*}

\subsection{Mixing and Quality Filtering}

For each target language $\ell$, we aggregate a set of publicly available pretraining corpora $\{\mathcal{D}_{\ell}^{(s)}\}_{s=1}^{S_\ell}$, which tend to be web-crawl derivatives released by different groups. We treat each corpus as an independent source and preserve source identity throughout processing. Each document is stored as a tuple $(x, s)$ where $x$ is the raw text and $s$ denotes the originating dataset. This bookkeeping is required for our downstream cross-source matching.

After aggregating, we apply quality filters tailed to each language, we do this to avoid common pitfalls in pre-existing large-scale multilingual filters. We therefore apply a language-specific set of filters $F_\ell(\cdot)$ to each source independently, producing
\[
\widetilde{\mathcal{D}}_{\ell}^{(s)} 
= \{(x, s) \in \mathcal{D}_{\ell}^{(s)} : F_\ell(x)=1\}~.
\]

While the exact specifications vary by language, these filters broadly fall into a small number of reusable categories:

\begin{itemize}[noitemsep]
    \item \textbf{Length and structure:} includes minimum document length and minimum number of tokens/words, rejection of documents dominated by short lines, and a minimum words-per-line constraint to remove boilerplate text.
    \item \textbf{Character and n-gram repetition:} constraints on excessive character repetitions and concentration of frequent n-grams that target spam and templated pages.
    \item \textbf{Code and template removal:} removal of documents dominated by JavaScript/code-like patterns, placeholder text, and policy/legal boilerplate.
    \item \textbf{Script and language consistency:} a minimum fraction of characters in the expected script (for example, Arabic script for Arabic, Latin for Turkish), along with a language identification score threshold.
\end{itemize}

We initialize thresholds from strong public baselines, such as FineWeb-2, but adapt them to language-specific writing conventions. For instance, we allow a zero terminal punctuation rate in Arabic (since it is common in informal web text), while enforcing stricter terminal punctuation and language identification thresholds in Latin script languages. This stage is intentionally lightweight; its role is to remove obvious low-quality content before similarity-based processing. 

\subsection{MinHash Deduplication}

After quality filtering, we perform near-duplicate removal jointly over the union of filtered sources $\widetilde{\mathcal{D}}_{\ell}=\bigcup_{s=1}^{S_\ell}\widetilde{\mathcal{D}}_{\ell}^{(s)}$. For each document $x$, we form a set of character $n$-gram shingles $\mathcal{S}(x)$, with $n=5$. We compute a MinHash signature of length $H=112$ (approximating Jaccard similarity between shingle sets), and apply banded Locality-Sensitive Hashing (LSH) with $B=14$ bands and $R=8$ hashes per band (so $H=B\cdot R=112$). Documents that collide in at least one band become candidate pairs. We then treat a candidate pair as a near-duplicate if its MinHash-estimated similarity exceeds a threshold $\tau=0.8$. We implement MinHash deduplication using standard large-scale components:
\begin{enumerate}[noitemsep]
    \item \textbf{Signature generation:} compute MinHash signatures for each $(x,s)\in \widetilde{\mathcal{D}}_\ell$ from its 5-gram shingle set.
    \item \textbf{Banding (LSH):} split each signature into $B$ bands of $R$ hashes and group documents by identical band keys to produce candidate pairs.
    \item \textbf{Clustering:} verify candidates with the MinHash similarity estimate and cluster all pairs with similarity $\ge \tau$ using Union-Find; each connected component is a near-duplicate cluster.
    \item \textbf{Filtering:} retain a single representative document per cluster and keep all non-duplicate documents.
\end{enumerate}

To ensure deterministic outputs, we select the representative as the element with minimum global processing index within each cluster (the first seen under a fixed traversal order). This makes deduplication reproducible across runs. We denote the resulting deduplicated set by $\mathcal{U}_\ell$.

\subsection{Cross-Source Matching as Ensemble Filtering}

Deduplication removes redundancy, but it also produces a valuable byproduct: clusters that span multiple source datasets. We leverage this cross-source agreement as a quality signal, treating each dataset's filtering pipeline as an independent ``vote'' on quality.

Consider each source dataset $\mathcal{D}^{(s)}$ as the output of an independent filtering process applied to a shared pool of web content. Each filtering pipeline makes different decisions: different crawl snapshots, different language identification thresholds, different quality heuristics. A document that survives multiple independent filters provides stronger evidence of quality than one retained by a single filter.

We rely on an assumption grounded in ensemble learning: that each pipeline's retention decisions are \emph{positively correlated} with some latent quality signal. Under this assumption, cross-source agreement reduces variance. A document retained by a single pipeline may have passed due to that pipeline's filtering choices or random chance. A document retained by multiple pipelines is less likely to represent a ``false positive'', content that slipped through due to filter-specific blind spots. This is analogous to inter-annotator agreement in crowdsourcing, where consensus among independent annotators indicates label reliability, even when no ground truth is available~\citep{dawid1979maximum,krippendorff2011computing}.

Formally, suppose each pipeline $s$ retains document $x$ with probability $q_s(x)$, where $q_s(x)$ is positively correlated with an unobserved quality variable $Q(x)$. Documents with high $|S(x)|$ (appearing in many sources) have, in expectation, higher $Q(x)$ than documents with low $|S(x)|$, not because any single pipeline is guaranteed to be accurate, but because agreement across pipelines with different error modes filters out noise. Let $\{C_i\}_{i=1}^m$ be the set of near-duplicate clusters (connected components) formed during MinHash deduplication over $\widetilde{\mathcal{D}}_\ell$. For a cluster $C_i$, define its set of unique sources as
\[
S(C_i)
=\{\, s : (x,s)\in C_i \,\}~.
\]
We define the matched subset as the representatives of clusters that appear in at least two sources:
\[
\mathcal{M}_\ell 
= \{ \operatorname{rep}(C_i) : |S(C_i)| \ge 2 \}~.
\]
In practice, we construct $\mathcal{M}_\ell$ by reusing the near-duplicate links from the MinHash-LSH pipeline and restricting to cross-source pairs ($s(u)\neq s(v)$), which reduces memory and avoids counting within-source redundancy as agreement. The source count for each document is preserved as metadata, enabling downstream filtering by agreement strength, for example requiring 3 or more sources instead of 2.

A key advantage of cross-source matching is that it requires no computation beyond standard deduplication. When combining multiple web corpora, MinHash deduplication is already necessary to avoid training on near-duplicate content~\citep{lee2022deduplicating}. Cross-source matching examines the source labels within clusters this pipeline already produces, propagating labels through Union-Find and counting unique sources per cluster are $O(1)$ operations. In contrast, model-based quality filtering requires inference over every document: for a 300B token corpus, even efficient scoring at 10,000 tokens/second requires ${\sim}8{,}000$ GPU-hours.

\begin{table}[t]
\centering
\caption{Token reduction through MixMinMatch pipeline stages. QF: Quality Filtered, MH: MinHash-deduplicated, MAT: Matched.}
\label{tab:reduction_ratios}
\small
\begin{tabular}{@{}l r r@{}}
\toprule
\textbf{Language} & \textbf{QF $\rightarrow$ MH} & \textbf{QF $\rightarrow$ MAT} \\
\midrule
Arabic  & 40.9\% & 82.0\% \\
Turkish & 45.4\% & 81.8\% \\
Hindi   & 41.5\% & 79.2\% \\
\midrule
\textbf{Average} & \textbf{39.8\%} & \textbf{82.8\%} \\
\bottomrule
\end{tabular}
\end{table}

\section{Web Scraping Redundancies}

We apply the MixMinMatch pipeline to Arabic, Turkish, and Hindi, producing language-specific datasets that we denote AraMix, TurMix, and HinMix, respectively. Alongside the Arabic sources, we include two language-specific sources: Sangraha~\citep{khan2024sangraha} for Hindi and VNGRS-Web~\citep{vngrs2024web} for Turkish. For each language, we release two subsets: the full MinHash deduplicated corpus and the smaller matched subset. We select Arabic, Turkish, and Hindi as our three languages for their script diversity.

\begin{table}[h]
\centering
\caption{Document survival rates after cross-dataset MinHash deduplication, computed from post-quality-filtering counts. Sources are grouped by language and sorted by survival rate.}
\label{tab:survival_all}
\small
\begin{tabular}{@{}llrrr@{}}
\toprule
\textbf{Language} & \textbf{Source} & \textbf{Before} & \textbf{After} & \textbf{Survival} \\
\midrule
Arabic
 & FinePDFs        &  0.68M &  0.67M & 98.5\% \\
 & ArabicWeb24     & 36.81M & 34.54M & 93.8\% \\
 & CulturaX        & 68.08M & 44.24M & 65.0\% \\
 & C4              & 44.91M & 26.86M & 59.8\% \\
 & FineWeb-2       & 57.20M & 32.05M & 56.0\% \\
 & ClusterLab      & 11.63M &  6.53M & 56.1\% \\
 & HPLT 2.0        & 64.05M & 34.01M & 53.1\% \\
\midrule
Hindi
 & FineWeb-2              & 21.79M & 17.15M & 78.7\% \\
 & CulturaX               & 19.21M & 11.51M & 59.9\% \\
 & Sangraha (U)  & 16.65M &  8.94M & 53.7\% \\
 & Sangraha (V)    & 17.06M &  9.05M & 53.1\% \\
 & HPLT 2.0               & 12.72M &  6.67M & 52.4\% \\
 & C4                     & 12.22M &  6.25M & 51.1\% \\
\midrule
Turkish
 & FineWeb-2  & 90.18M & 54.53M & 60.5\% \\
 & CulturaX   & 86.40M & 47.87M & 55.4\% \\
 & HPLT 2.0   & 99.10M & 53.71M & 54.2\% \\
 & VNGRS-Web  & 49.36M & 26.50M & 53.7\% \\
 & C4         & 68.97M & 36.46M & 52.9\% \\
\bottomrule
\end{tabular}
\end{table}

The survival rate column in Table~\ref{tab:survival_all} quantifies how much unique content each source contributes after cross-dataset deduplication. This metric is defined as the fraction of documents retained after removing cross-source duplicates. In Arabic, FinePDFs retains 98.5\% of its content, reflecting the unique PDF-derived sources used, FineWeb-2 only retains 56\%. ArabicWeb24 uniquely achives a 93.8\% survival rate, suggesting its processing pipeline is able to capture relevant high-quality data from Common Crawl snapshots that is typically dropped by other pipelines. These statistics carry practical implications. Sources with high survival rates represent sources that either sought out unique sources to use or substantially modified their processing pipeline to capture typically dropped content. Complete token and document breakdowns by source for both matched and MinHash corpora appear in Appendix~\ref{sec:sources_appendix}. We further characterize the domain composition of AraMix-MinHash in Appendix~\ref{sec:domain_appendix}.

These statistics surface an apparent tension: ArabicWeb24 achieves 93.8\% survival after cross-dataset deduplication, indicating highly unique content, yet it is also the strongest single-source baseline. If cross-source agreement were the sole indicator of quality, ArabicWeb24's unique content should underperform, but it does not.

This tension dissolves once we recognize that cross-source agreement is a \emph{sufficient} but not \emph{necessary} signal of quality. Content retained by multiple independent pipelines is reliably high-quality, but content retained by only one pipeline is not necessarily low-quality,particularly when that pipeline employs sophisticated language-specific filtering. Our experimental results support this interpretation. AraMix-Matched outperforms ArabicWeb24 (0.161 vs.\ 0.154), demonstrating that cross-source agreement identifies additional high-quality content beyond what any single source provides. AraMix-Matched \emph{without} ArabicWeb24 still matches the ArabicWeb24 baseline, indicating that cross-source agreement among the remaining generic pipelines identifies content of comparable quality to a purpose-built corpus. The two signals are complementary: language-specific curation captures valuable content that cross-source agreement misses, while cross-source agreement validates content across sources. MixMinMatch benefits from both. This complementarity suggests a natural extension: rather than treating all sources equally, a weighted voting scheme could assign higher weight to sources with demonstrated quality, for instance, upweighting ArabicWeb24's vote based on its downstream performance or survival rate. We leave exploration of such weighted ensembles to future work, noting that even uniform weighting already yields substantial gains.

\section{Evaluation Protocol}

We evaluate the performance of the generated pretraining datasets by using the FineWeb-2 setup. That is, we measure the performance of downstream tasks of an LLM pretrained on a fixed token budget using various datasets. By following the FineWeb-2 setup, we can have a direct comparison between our dataset variants and the strongest baselines reported in that work.

\subsection{Datasets}

For each language $\ell$, we begin with a set of sources $\{\mathcal{D}_{\ell}^{(s)}\}_{s=1}^{S_\ell}$, apply the language-specific quality filters, and run cross-source MinHash deduplication over the union. Deduplication groups near-duplicate documents into clusters and keeps one deterministic representation per cluster, producing the MinHash deduplicated dataset (labeled MinHash). In addition, each representative sample carries a source count: the number of unique sources that contributed a near-duplicate to its cluster. Intuitively, a higher source count indicates that multiple independent dataset sources recovered the same underlying content, our core quality signal. We use this source count to define the matched subset and two ablations variants remove baseline influence.

\subsubsection{Corpus Variants}

We evaluate five corpus configurations for each language to isolate the contribution of each MixMinMatch component. Let $b$ denote the best-performing single-source baseline for language $\ell$, as identified in the FineWeb-2 report.

\begin{itemize}[noitemsep]
    \item \textbf{Baseline:} The source $b$ alone, under the same token budget as all other runs.
    \item \textbf{MixMinMatch-MinHash:} The full MinHash deduplicated mixture over all sources.
    \item \textbf{MixMinMatch-Matched:} The subset of MinHash representatives whose clusters span at least two sources. This retains only content validated by cross-source agreement.
    \item \textbf{MinHash without baseline:} The MinHash mixture with baseline only documents removed. We drop representatives originating from $b$ if their cluster has only one source. This tests whether gains derive primarily from content unique to the best baseline.
    \item \textbf{Matched without baseline:} The matched subset with baseline influence removed. For clusters including $b$, we require at least two additional source counts. This ensures observed quality cannot be attributed to the baseline alone.
\end{itemize}

\subsubsection{Proportional Sampling to a Fixed Token Budget}

All multi-source pools are larger than our training budget. To keep comparisons fair, we sample each training pool to the same token budget while preserving its natural source mixture. For a given subset, we compute the fraction of available documents contributed by each source. We then allocate tokens to sources according to their document fractions. Finally, we uniformly sample the required number of tokens per source, concatenate, and shuffle to produce the training dataset. This proportional procedure is applied to the MinHash pool, the matched pool, and both baseline ablated pools. This ensures that the pools we generate are representative of the datasets as a whole.

\subsection{Training}
\label{sec:training_setup}


We train Llama-style decoder-only models, using \texttt{nanotron}, with 1.46 billion parameters, following the same optimization recipe across all experiments. The model uses 14 layers, hidden size of 2048, intermediate size of 8192, and 32 attention heads with a sequence length of 2048. We tokenize with \texttt{google/gemma-2b}'s tokenizer~\citep{gemmateam2024gemma2improvingopen} and use a vocabulary size of 256k. Details on the optimizer are present in Table~\ref{tab:training_hparams}.

\begin{table}[h]
\centering
\caption{Training hyperparameters and token budget.}
\label{tab:training_hparams}
\small
\begin{tabular}{@{}l l@{}}
\toprule
\textbf{Setting} & \textbf{Value} \\
\midrule
Optimizer & Fused AdamW \\
Peak learning rate & $3\times 10^{-4}$ \\
Adam $\beta_1$, $\beta_2$ & $0.9$, $0.95$ \\
Weight decay & $0.1$ \\
Warmup & Linear warmup 500 steps  \\
Decay & Cosine decay 13,500 steps \\
Minimum learning rate & $3\times 10^{-5}$ \\
Training steps & 14{,}000 \\
Micro-batch size (per replica) & 4 \\
Gradient accumulation & 32 \\
Data parallelism (DP) & 8 \\
Global batch (sequences/step) & $1024$ \\
Tokens per step & $2{,}097{,}152$ \\
Total training tokens & $29.36$B \\
\bottomrule
\end{tabular}
\end{table}

\subsection{Evaluation}
\label{evaluation_section}

We assess the quality of our models by closely following the FineWeb-2 recipe. This involves selecting evaluation tasks that show monotonicity, relatively low-noise, non-random performance, and model ordering consistency. The full list of tasks chosen per language with complete benchmark descriptions and random baselines is reported in Appendix~\ref{sec:eval_suite_appendix}.

\subsubsection{Score Normalization}

Raw benchmark scores are not directly comparable across tasks due to varying numbers of answer choices in multiple-choice formats. We normalize scores to account for random baseline performance. For a given task, the rescaled score is computed as
\[
s_{\text{rescaled}} 
= \frac{s_{\text{raw}} - b}{1 - b},
\]
where $s_{\text{raw}}$ denotes the raw accuracy and $b$ represents the random baseline corresponding to uniform random guessing. Table~\ref{tab:baselines} summarizes the baseline values by task format.

\subsubsection{Aggregation Procedure}

The aggregate score is computed via a two-step hierarchical averaging procedure. First, rescaled scores are averaged within each evaluation category. Second, a macro-average is computed across all categories, weighting each category equally regardless of the number of constituent benchmarks. Formally, let $\mathcal{C}$ denote the set of evaluation categories and let $\mathcal{T}_c$ denote the set of tasks within category $c \in \mathcal{C}$. The aggregate score is given by
\[
    s_{\text{agg}} = \frac{1}{|\mathcal{C}|} \sum_{c \in \mathcal{C}} \left( \frac{1}{|\mathcal{T}_c|} \sum_{t \in \mathcal{T}_c} s_{\text{rescaled}}^{(t)} \right)~.
\]
This formulation prevents categories with more benchmarks from dominating the aggregate score.

\section{Results}

\begin{figure*}[!t]
    \centering

    \begin{minipage}{\textwidth}
        \centering
        \begin{minipage}[t]{0.33\textwidth}\centering
            \includegraphics[width=\linewidth]{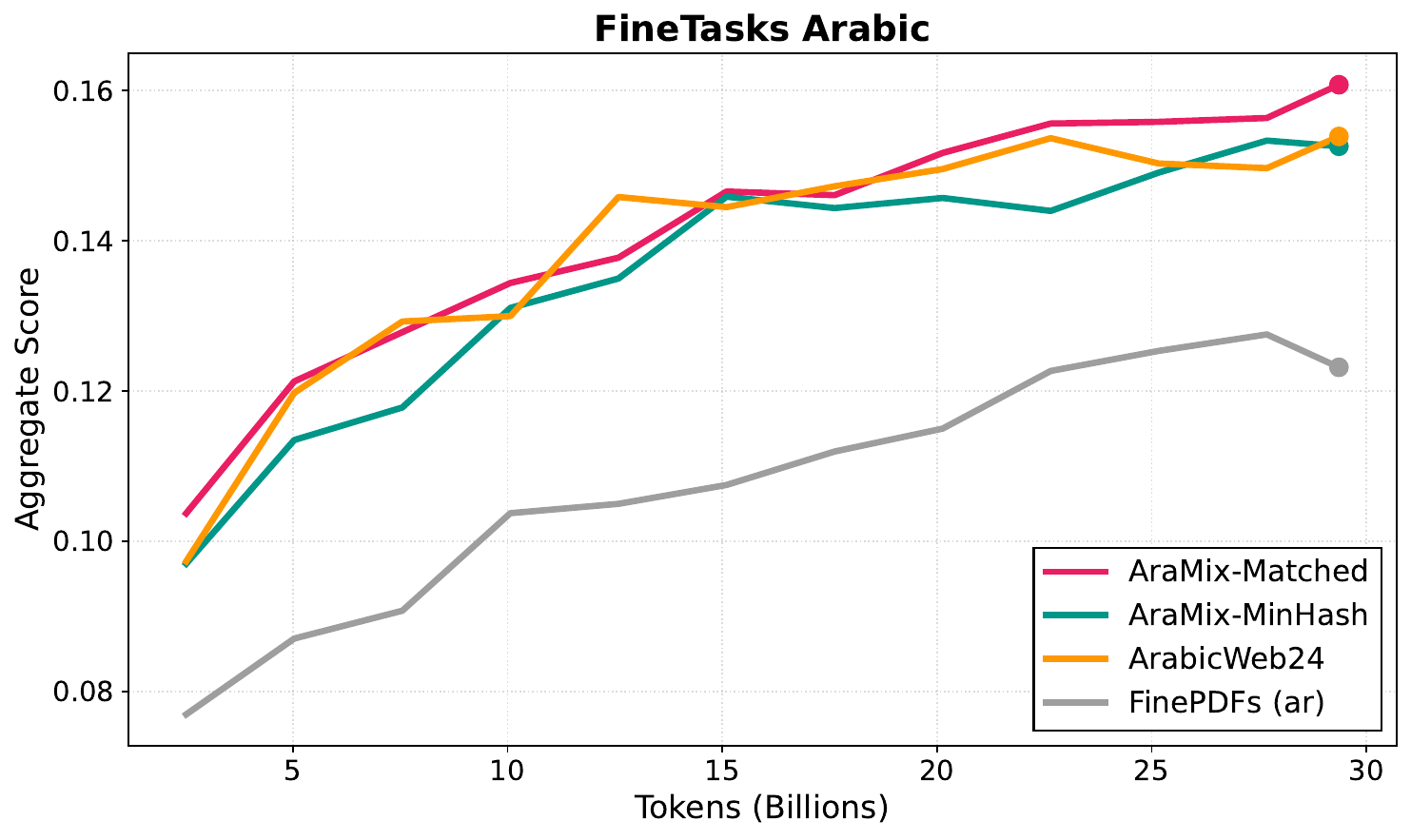}
        \end{minipage}\hfill
        \begin{minipage}[t]{0.33\textwidth}\centering
            \includegraphics[width=\linewidth]{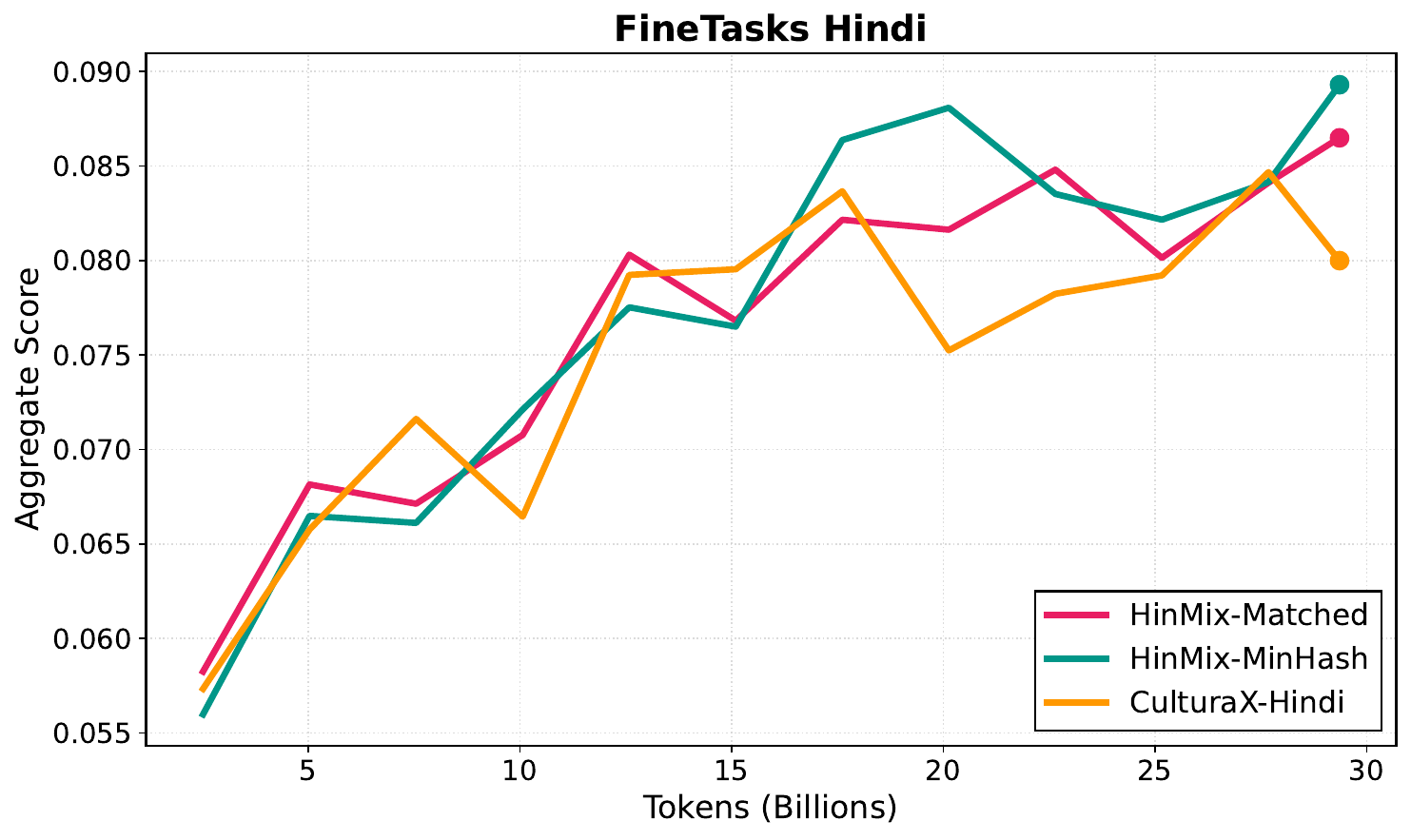}
        \end{minipage}\hfill
        \begin{minipage}[t]{0.33\textwidth}\centering
            \includegraphics[width=\linewidth]{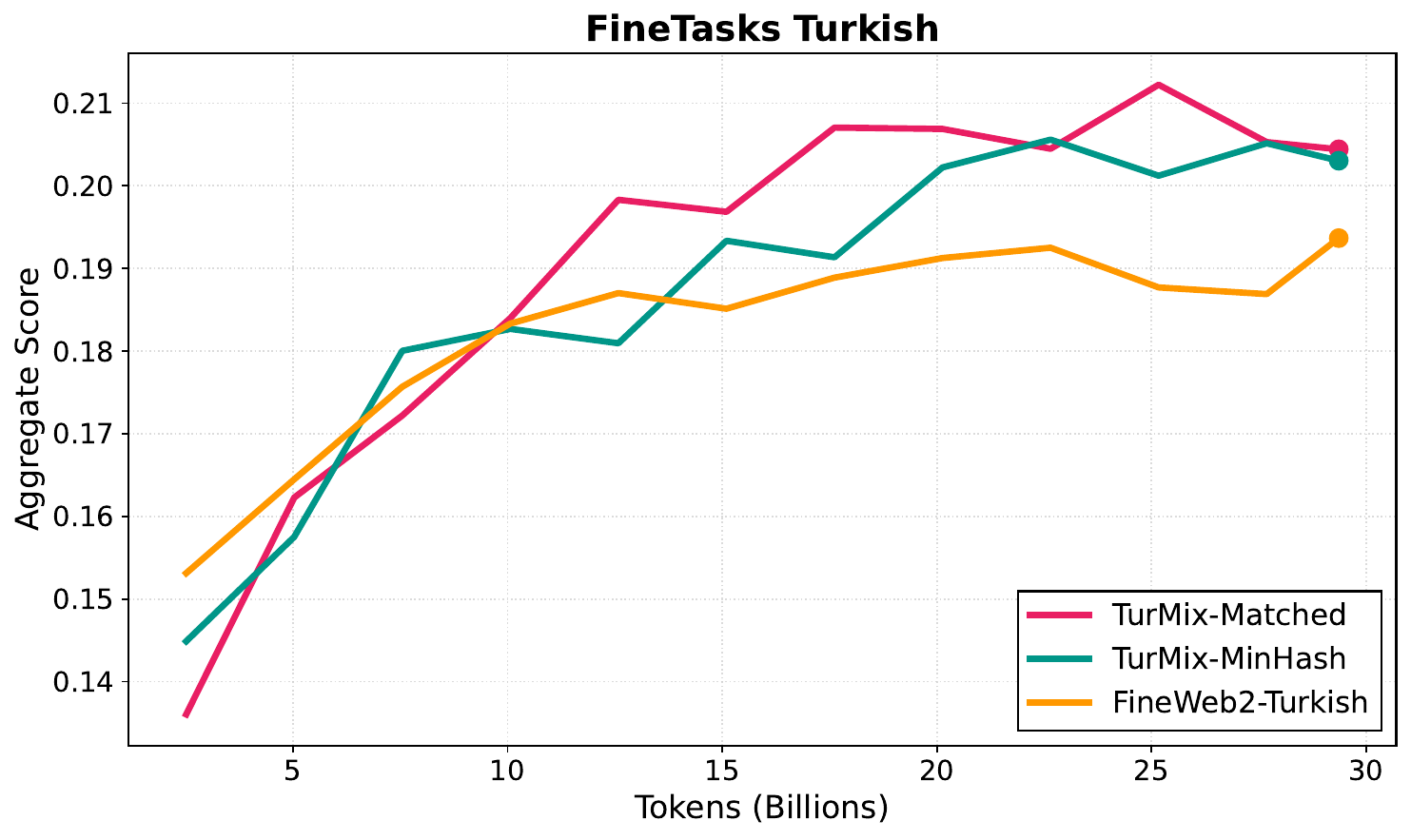}
        \end{minipage}

        \captionof{figure}{Aggregate FineTasks score of Arabic, Hindi, and Turkish trained models at different checkpoints throughout 30 billion token training runs.}
        \label{fig:benchmark-figs-with-baseline}
    \end{minipage}

    \vspace{0.6cm}

    \begin{minipage}{\textwidth}
        \centering
        \begin{minipage}[t]{0.33\textwidth}\centering
            \includegraphics[width=\linewidth]{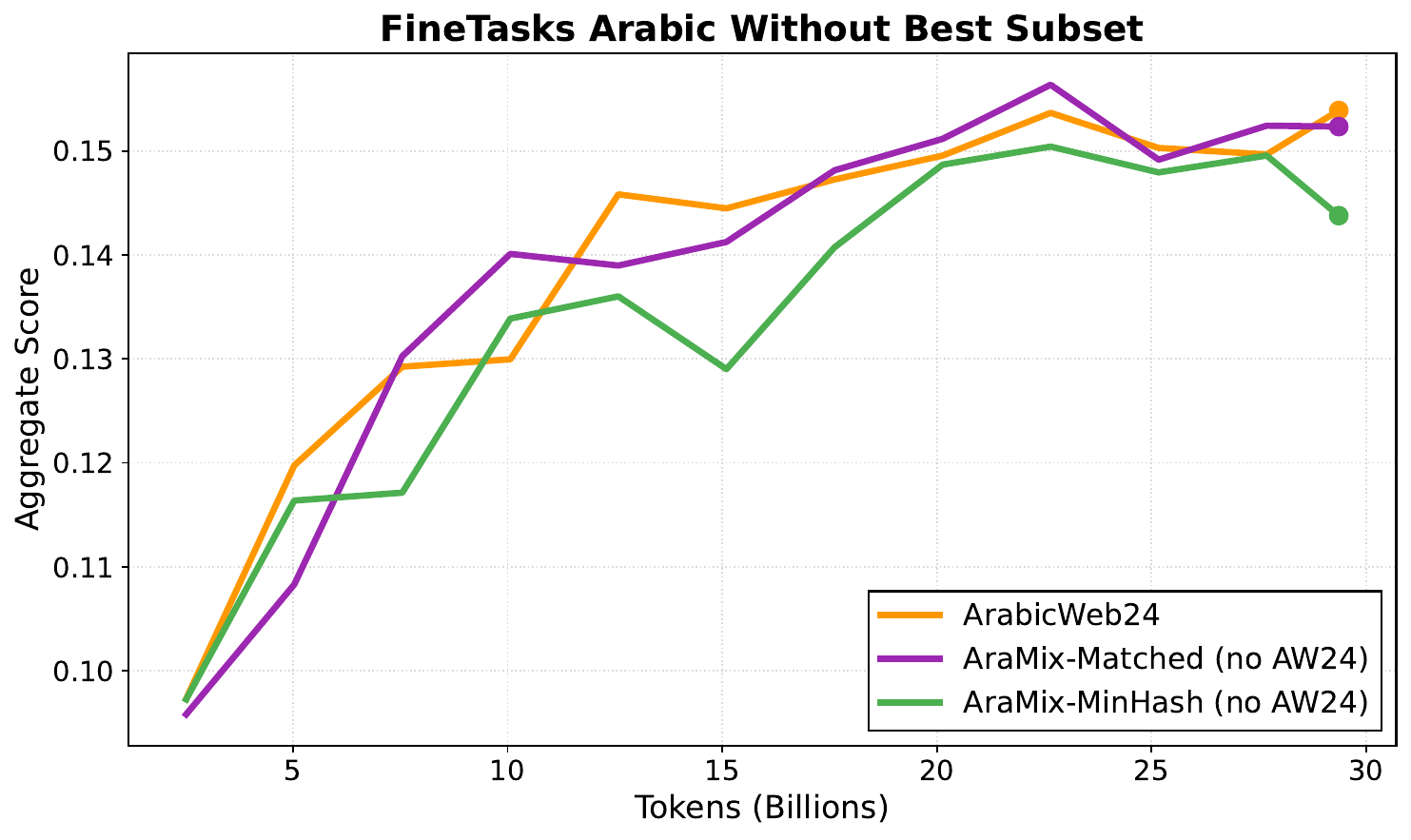}
        \end{minipage}\hfill
        \begin{minipage}[t]{0.33\textwidth}\centering
            \includegraphics[width=\linewidth]{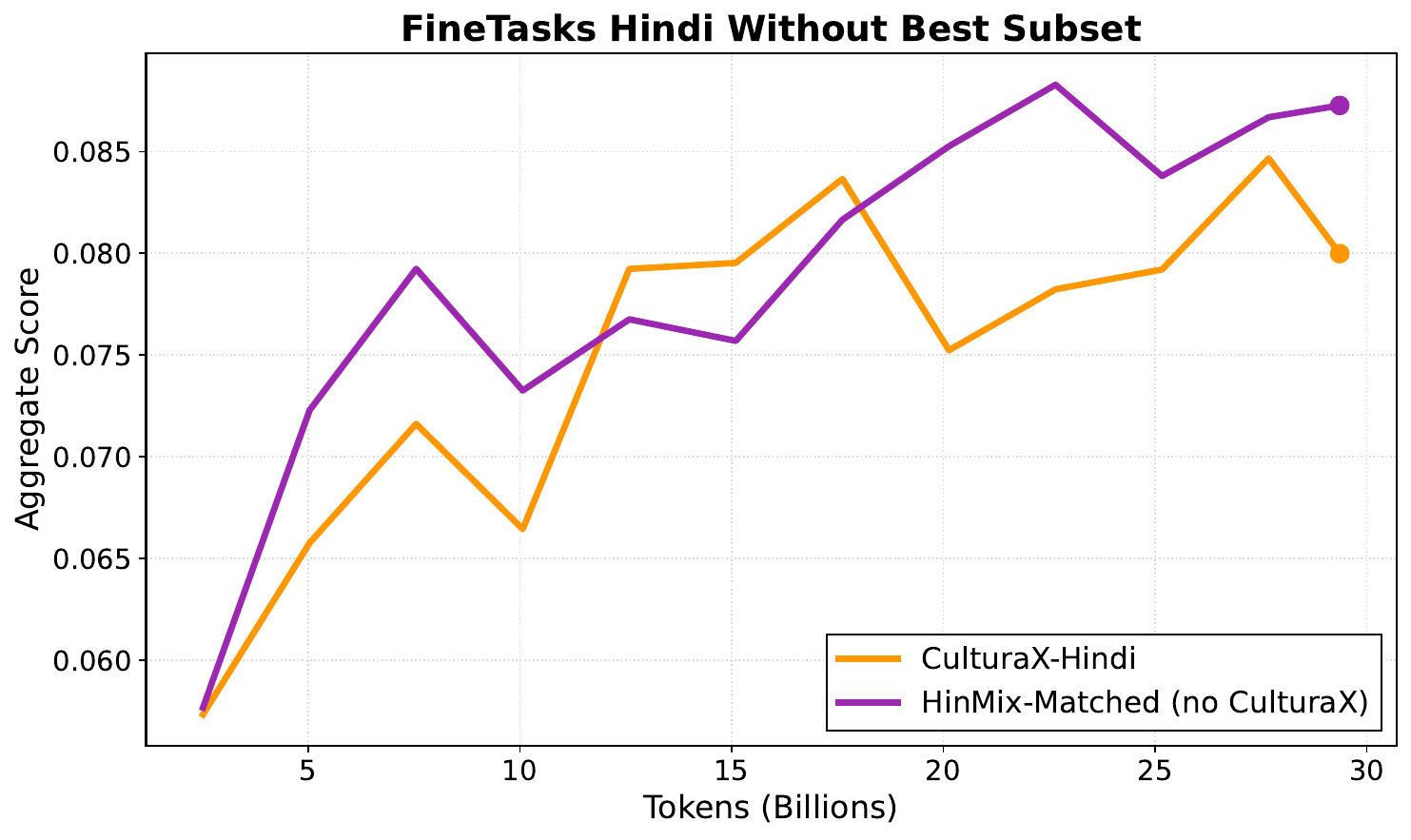}
        \end{minipage}\hfill
        \begin{minipage}[t]{0.33\textwidth}\centering
            \includegraphics[width=\linewidth]{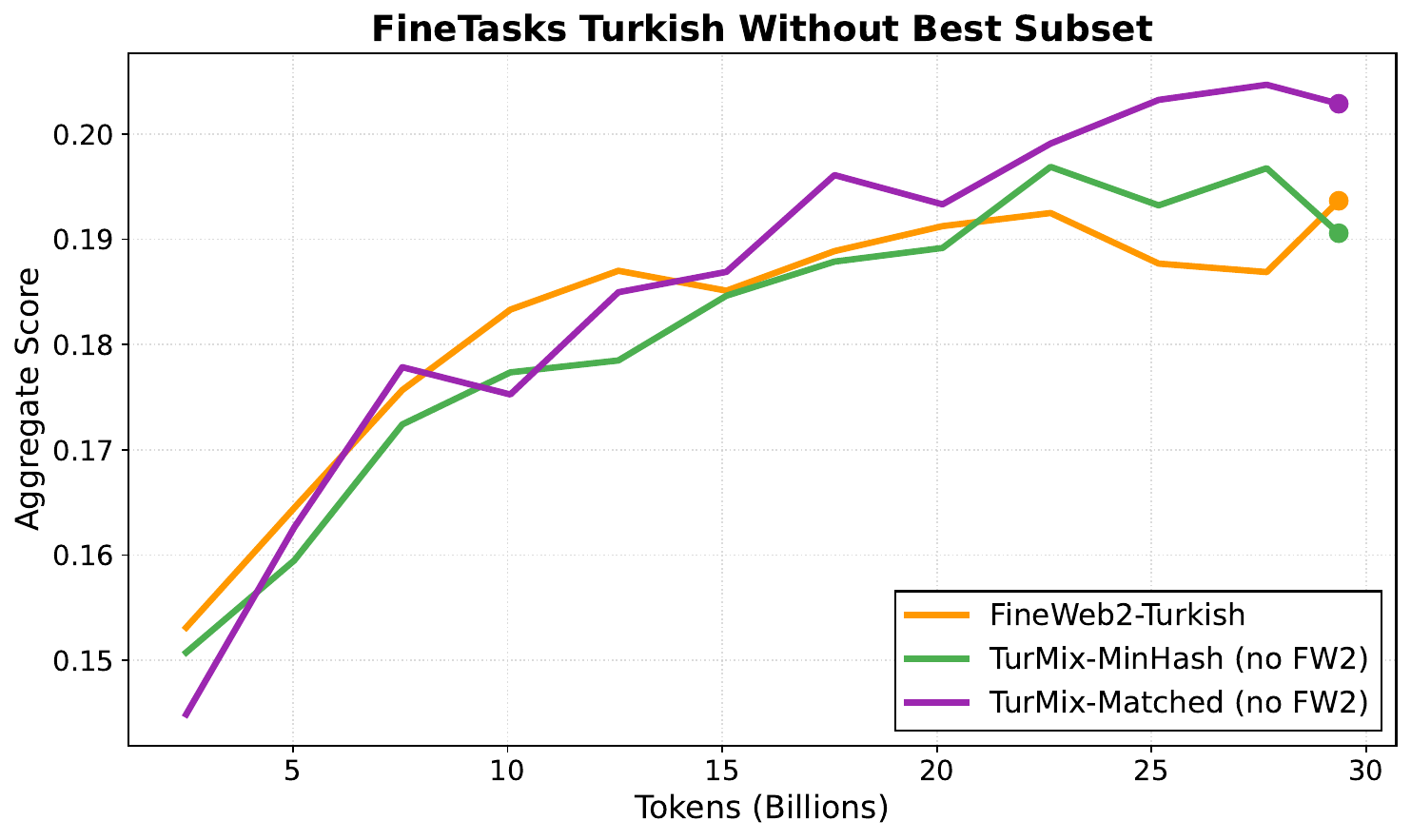}
        \end{minipage}

        \captionof{figure}{Aggregate FineTasks score of Arabic, Hindi, and Turkish trained models at different checkpoints, with the best performing subset of each MixMinMatch dataset removed.}
        \label{fig:benchmark-figs-without-baseline}
    \end{minipage}
\end{figure*}

We evaluate MixMinMatch through the training procedure outlined in Section~\ref{sec:training_setup} on each corpus variant and measure aggregate FineTasks scores as described in Section~\ref{evaluation_section}. Figures~\ref{fig:benchmark-figs-with-baseline} and \ref{fig:benchmark-figs-without-baseline} present the FineTask score of each variant.

\subsection{MixMinMatch Exceeds Single-Source Baselines}

Figure~\ref{fig:benchmark-figs-with-baseline} compares each MixMinMatch variant to the best-performing single-source baseline for that language. Our matched subsets consistently outperform or match baselines, while our MinHash deduplicated mixtures provide substantially larger corpora at competitive quality, most times outperforming the baseline as well.

\begin{table}[t]
\centering
\caption{Dataset sizes (in billions of tokens) for the single-source baseline, MinHash mixture, and matched subset in billions of tokens.}
\label{tab:dataset_size_comparison}
\small
\begin{tabular}{@{}l r r r@{}}
\toprule
\textbf{Language} & \textbf{Single-source} & \textbf{MinHash} & \textbf{Matched} \\
\midrule
Arabic  & 41B  & 178B & 54B \\
Hindi   & 29B & 76B & 27B \\
Turkish & 79B & 168B & 56B \\
\bottomrule
\end{tabular}
\end{table}

\noindent\textbf{Arabic.} AraMix-Matched achieves an aggregate score of 0.161, compared to 0.154 for ArabicWeb24---a 4.5\% relative improvement. This gain is consistent across checkpoints: the matched subset outperforms or matches the baseline at every evaluation point after 10B tokens, suggesting the improvement reflects quality differences rather than noise. AraMix-MinHash achieves 0.153, matching the baseline while providing 4.3$\times$ more tokens (178B vs.\ 41B, Table~\ref{tab:dataset_size_comparison}). For practitioners with sufficient compute, the MinHash mixture offers a substantially larger high-quality corpus at no quality penalty.

\noindent\textbf{Hindi.} HinMix-MinHash achieves 0.089, improving over CulturaX-Hindi (0.080) by 11.6\%. the largest relative gain across our experiments. HinMix-Matched scores 0.087, also substantially outperforming the baseline. Both variants show consistent improvement across training checkpoints.

\noindent\textbf{Turkish.} TurMix-Matched achieves 0.204, surpassing FineWeb-2's Turkish subset (0.194) by 5.2\%. TurMix-MinHash scores 0.203, surpassing FineWeb-2's Turkish subset while providing more data.

\subsection{Gains Persist Beyond Single Source Removal}

\begin{table}[t]
\centering
\caption{Final aggregate FineTasks scores. Relative improvement computed against single-source baseline. Ablations (marked $\dagger$) exclude content unique to the baseline source.}
\label{tab:main_results}
\small
\begin{tabular}{@{}l l r r@{}}
\toprule
\textbf{Lang.} & \textbf{Corpus} & \textbf{Score} & \textbf{$\Delta$} \\
\midrule
\multirow{5}{*}{Arabic}
 & ArabicWeb24 (baseline) & 0.154 & --- \\
 & AraMix-MinHash & 0.153 & $-$0.6\% \\
 & AraMix-Matched & \textbf{0.161} & $+$4.5\% \\
 & AraMix-MinHash$^\dagger$ & 0.144 & $-$6.5\% \\
 & AraMix-Matched$^\dagger$ & 0.152 & $-$1.3\% \\
\midrule
\multirow{4}{*}{Hindi}
 & CulturaX (baseline) & 0.080 & --- \\
 & HinMix-MinHash & \textbf{0.089} & $+$11.6\% \\
 & HinMix-Matched & 0.087 & $+$8.1\% \\
 & HinMix-Matched$^\dagger$ & 0.087 & $+$9.1\% \\
\midrule
\multirow{5}{*}{Turkish}
 & FineWeb-2 (baseline) & 0.194 & --- \\
 & TurMix-MinHash & 0.203 & $+$4.8\% \\
 & TurMix-Matched & \textbf{0.204} & $+$5.5\% \\
 & TurMix-MinHash$^\dagger$ & 0.191 & $-$1.6\% \\
 & TurMix-Matched$^\dagger$ & 0.203 & $+$4.7\% \\
\bottomrule
\end{tabular}
\end{table}

A natural concern is whether MixMinMatch inherits quality from the best baseline source. Figure~\ref{fig:benchmark-figs-without-baseline} addresses this by removing documents that appear only in the baseline source.

\noindent\textbf{Arabic.} AraMix-Matched without ArabicWeb24 scores 0.152, performing on-par with the ArabicWeb24 baseline. Even after removing all ArabicWeb24-only content, the cross-source signal among remaining sources identifies a corpus nearly as good as the best single source. The MinHash variant without ArabicWeb24 drops to 0.144, a more substantial degradation, suggesting that ArabicWeb24's unique content contributes meaningfully to the full mixture's quality. The lack of datasets to `corroborate' ArabicWeb24's samples presents itself as a weakness here. This opens up future work to explore a weighted voting mechanism.

\noindent\textbf{Hindi.} HinMix-Matched without CulturaX achieves 0.087, matching the full matched subset and \emph{outperforming} the CulturaX baseline. This provides strong evidence that cross-source agreement among sources other than CulturaX identifies high-quality content independently of any single corpus.

\noindent\textbf{Turkish.} TurMix-Matched without FineWeb-2 achieves 0.203, outperforming the FineWeb-2 baseline despite excluding all FineWeb-2-only content. In contrast, TurMix-MinHash without FineWeb-2 drops to 0.191, below the baseline. This asymmetry suggests that cross-source agreement captures quality that the full mixture partially obscure.

\subsection{Comparison with Model-Based Quality Filtering}


\begin{figure}
    \centering
    \includegraphics[width=0.6\linewidth]{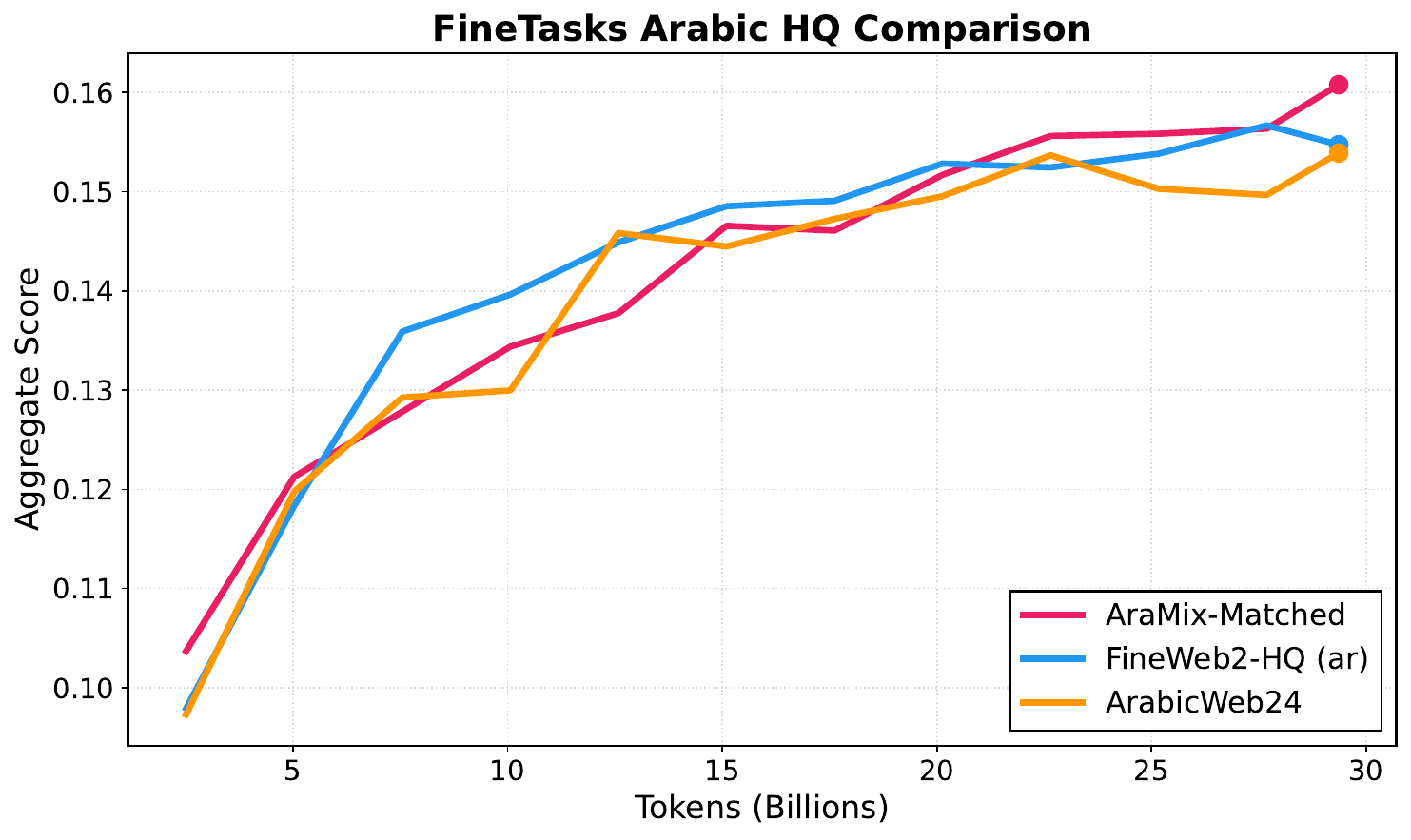}
    \caption{The matched subset of AraMix outperforms model-based scoring and filtering of pretraining corpora.}
    \label{fig:matched_against_model_based}
\end{figure}

Figure~\ref{fig:matched_against_model_based} compares AraMix-Matched against FineWeb2-HQ~\citep{messmer2025enhancingmultilingualllmpretraining}, a corpus filtered using model-based quality scoring. \emph{The matched subset outperforms the model-scored dataset despite being model-free.} This comparison highlights the practical advantage of cross-source matching: model-based filtering requires substantial compute for inference (proportional to corpus size), while cross-source matching extracts a quality signal from deduplication infrastructure that practitioners already deploy. However, we emphasize that these approaches are complementary. Cross-source matching could serve as a first-stage filter, reducing corpus size before applying more expensive model-based refinement. We leave exploration of such hybrid pipelines to future work.

\section{Conclusion}

We have presented MixMinMatch, a method for building multilingual pretraining corpora by combining existing datasets and using cross-source agreement as a quality signal. Our key insight is that content independently retained by multiple filtering pipelines is more likely to be high-quality, and that this signal can be extracted for free from standard MinHash deduplication infrastructure. We discuss limitations and potential future work in Appendix~\ref{sef:limitations_appendix}.

Applied to Arabic, Turkish, and Hindi, MixMinMatch produces corpora that match or exceed the quality of the best single-source baselines. The matched subsets achieve 4--11\% relative improvements while often using fewer tokens than baselines. Ablations confirm that these gains do not merely inherit quality from the strongest single source: cross-source agreement captures quality that no individual corpus provides alone.

Beyond the immediate empirical results, our work suggests a broader principle: \emph{the redundancy in existing web corpora is a resource, not just waste}. Multiple groups have invested substantial effort in crawling and filtering the web for different languages. Rather than each new project starting from scratch, the research community can extract additional value by systematically combining what already exists.

\section*{Acknowledgements}

The research reported in this publication was supported by funding from King Abdullah University of Science and Technology (KAUST) - Center of Excellence for Generative AI, under award number 5940.

We also thank our lab member, Manar Alnazer, for her work on filtering datasets from Common Crawl through \texttt{datatrove}. The work helped deepen our understanding of how publicly available dataset filters are designed.

\bibliography{custom}

@article{raffel2020exploring,
  title   = {Exploring the Limits of Transfer Learning with a Unified Text-to-Text Transformer},
  author  = {Raffel, Colin and Shazeer, Noam and Roberts, Adam and Lee, Katherine and Narang, Sharan and Matena, Michael and Zhou, Yanqi and Li, Wei and Liu, Peter J.},
  journal = {Journal of Machine Learning Research},
  volume  = {21},
  number  = {140},
  pages   = {1--67},
  year    = {2020}
}

@inproceedings{penedo2024fineweb,
  title     = {The {FineWeb} Datasets: Decanting the Web for the Finest Text Data at Scale},
  author    = {Penedo, Guilherme and Kydlíček, Hynek and Ben Allal, Loubna and Lozhkov, Anton and Mitchell, Margaret and Raffel, Colin and Von Werra, Leandro and Wolf, Thomas},
  booktitle = {Advances in Neural Information Processing Systems},
  year      = {2024}
}

@article{penedo2025fineweb2,
  title = {{FineWeb2}: One Pipeline to Scale Them All - Adapting Pre-Training Data Processing to Every Language},
  author  = {Penedo, Guilherme and Kydlícek, Hynek and Sabolcec, Vinko and Messmer, Bettina and Foroutan, Negar and Kargaran, Amir Hossein and Raffel, Colin and Jaggi, Martin and von Werra, Leandro and Wolf, Thomas},
  journal = {arXiv preprint arXiv:2506.20920},
  year    = {2025}
}

@inproceedings{nguyen2024culturax,
  title     = {CulturaX: A Cleaned, Enormous, and Multilingual Dataset for Large Language Models in 167 Languages},
  author    = {Nguyen, Thu{\d{a}}t and Nguyen, Chien Van and Lai, Viet Dac and Man, Hieu and Ngo, Nghia Trung and Dernoncourt, Franck and Rossi, Ryan A. and Nguyen, Thien Huu},
  booktitle = {Proceedings of the 2024 Joint International Conference on Computational Linguistics, Language Resources and Evaluation (LREC-COLING)},
  year      = {2024},
  url       = {https://aclanthology.org/2024.lrec-main.377/}
}

@misc{farhat2024arabicweb24,
  title        = {{ArabicWeb24}: Creating a High Quality {Arabic} Web-Only Pre-Training Dataset},
  author       = {Farhat, May and Taghadouini, Said and Hallström, Oskar and Hajri-Gabouj, Sonja},
  year         = {2024},
  howpublished = {LightOn Blog},
  url          = {https://www.lighton.ai/lighton-blogs/arabicweb24}
}

@article{burchell2025hpltv2,
  title   = {An Expanded Massive Multilingual Dataset for High-Performance Language Technologies {(HPLT)}},
  author  = {Burchell, Laurie and de Gibert, Ona and Arefyev, Nikolay and Aulamo, Mikko and Ba{\~n}{\'o}n, Marta and Chen, Pinzhen and Fedorova, Mariia and Guillou, Liane and Haddow, Barry and Haji{\v{c}}, Jan and Helcl, Jind{\v{r}}ich and Henriksson, Erik and Klimaszewski, Mateusz and Komulainen, Ville and Kutuzov, Andrey and Kyt{\"o}niemi, Joona and Laippala, Veronika and M{\ae}hlum, Petter and Malik, Bhavitvya and Mehryary, Farrokh and Mikhailov, Vladislav and Moghe, Nikita and Myntti, Amanda and O'Brien, Dayy{\'a}n and Oepen, Stephan and Pal, Proyag and Piha, Jousia and Pyysalo, Sampo and Ram{\'\i}rez-S{\'a}nchez, Gema and Samuel, David and Stepachev, Pavel and Tiedemann, J{\"o}rg and Vari{\v{s}}, Du{\v{s}}an and Vojt{\v{e}}chov{\'a}, Tereza and Zaragoza-Bernabeu, Jaume},
  journal = {arXiv preprint arXiv:2503.10267},
  year    = {2025}
}

@misc{huggingfacefw2025finepdfs,
  title        = {{FinePDFs} Dataset Card},
  author       = {{Hugging Face}},
  year         = {2025},
  howpublished = {Hugging Face Datasets},
  url          = {https://huggingface.co/datasets/HuggingFaceFW/finepdfs}
}

@article{aloui2024arabic101b,
  title   = {101 Billion {Arabic} Words Dataset},
  author  = {Aloui, Manel and Chouikhi, Hasna and Chaabane, Ghaith and Kchaou, Haithem and Dhaouadi, Chehir},
  journal = {arXiv preprint arXiv:2405.01590},
  year    = {2024}
}

@article{rae2021gopher,
  title   = {Scaling Language Models: Methods, Analysis \& Insights from Training {Gopher}},
  author = {{Rae}, Jack W. and {Borgeaud}, Sebastian and {Cai}, Trevor and {Millican}, Katie and {Hoffmann}, Jordan and {Song}, Francis and {Aslanides}, John and {Henderson}, Sarah and {Ring}, Roman and {Young}, Susannah and {Rutherford}, Eliza and {Hennigan}, Tom and {Menick}, Jacob and {Cassirer}, Albin and {Powell}, Richard and {van den Driessche}, George and {Hendricks}, Lisa Anne and {Rauh}, Maribeth and {Huang}, Po-Sen and {Glaese}, Amelia and {Welbl}, Johannes and {Dathathri}, Sumanth and {Huang}, Saffron and {Uesato}, Jonathan and {Mellor}, John and {Higgins}, Irina and {Creswell}, Antonia and {McAleese}, Nat and {Wu}, Amy and {Elsen}, Erich and {Jayakumar}, Siddhant and {Buchatskaya}, Elena and {Budden}, David and {Sutherland}, Esme and {Simonyan}, Karen and {Paganini}, Michela and {Sifre}, Laurent and {Martens}, Lena and {Li}, Xiang Lorraine and {Kuncoro}, Adhiguna and {Nematzadeh}, Aida and {Gribovskaya}, Elena and {Donato}, Domenic and {Lazaridou}, Angeliki and {Mensch}, Arthur and {Lespiau}, Jean-Baptiste and {Tsimpoukelli}, Maria and {Grigorev}, Nikolai and {Fritz}, Doug and {Sottiaux}, Thibault and {Pajarskas}, Mantas and {Pohlen}, Toby and {Gong}, Zhitao and {Toyama}, Daniel and {de Masson d'Autume}, Cyprien and {Li}, Yujia and {Terzi}, Tayfun and {Mikulik}, Vladimir and {Babuschkin}, Igor and {Clark}, Aidan and {de Las Casas}, Diego and {Guy}, Aurelia and {Jones}, Chris and {Bradbury}, James and {Johnson}, Matthew and {Hechtman}, Blake and {Weidinger}, Laura and {Gabriel}, Iason and {Isaac}, William and {Lockhart}, Ed and {Osindero}, Simon and {Rimell}, Laura and {Dyer}, Chris and {Vinyals}, Oriol and {Ayoub}, Kareem and {Stanway}, Jeff and {Bennett}, Lorrayne and {Hassabis}, Demis and {Kavukcuoglu}, Koray and {Irving}, Geoffrey},
  journal = {arXiv preprint arXiv:2112.11446},
  year    = {2021}
}

@article{grattafiori2024llama3,
  author = {Dubey, Abhimanyu and Jauhri, Abhinav and Pandey, Abhinav and Kadian, Abhishek and Al-Dahle, Ahmad and Letman, Aiesha and Mathur, Akhil and Schelten, Alan and Yang, Amy and Fan, Angela and Goyal, Anirudh and Hartshorn, Anthony and Yang, Aobo and Mitra, Archi and Sravankumar, Archie and Korenev, Artem and Hinsvark, Arthur and Rao, Arun and Zhang, Aston and Rodriguez, Aurelien and Gregerson, Austen and Spataru, Ava and Roziere, Baptiste and Biron, Bethany and Tang, Binh and Chern, Bobbie and Caucheteux, Charlotte and Nayak, Chaya and Bi, Chloe and Marra, Chris and McConnell, Chris and Keller, Christian and Touret, Christophe and Wu, Chunyang and Wong, Corinne and Ferrer, Cristian Canton and Nikolaidis, Cyrus and Allonsius, Damien and Song, Daniel and Pintz, Danielle and Livshits, Danny and Esiobu, David and Choudhary, Dhruv and Mahajan, Dhruv and Garcia-Olano, Diego and Perino, Diego and Hupkes, Dieuwke and Lakomkin, Egor and AlBadawy, Ehab and Lobanova, Elina and Dinan, Emily and Smith, Eric Michael and Radenovic, Filip and Zhang, Frank and Synnaeve, Gabriel and Lee, Gabrielle and Anderson, Georgia Lewis and Nail, Graeme and Mialon, Gregoire and Pang, Guan and Cucurell, Guillem and Nguyen, Hailey and Korevaar, Hannah and Xu, Hu and Touvron, Hugo and Zarov, Iliyan and Ibarra, Imanol Arrieta and Kloumann, Isabel and Misra, Ishan and Evtimov, Ivan and Copet, Jade and Lee, Jaewon and Geffert, Jan and Vranes, Jana and Park, Jason and Mahadeokar, Jay and Shah, Jeet and van der Linde, Jelmer and Billock, Jennifer and Hong, Jenny and Lee, Jenya and Fu, Jeremy and Chi, Jianfeng and Huang, Jianyu and Liu, Jiawen and Wang, Jie and Yu, Jiecao and Bitton, Joanna and Spisak, Joe and Park, Jongsoo and Rocca, Joseph and Johnstun, Joshua and Saxe, Joshua and Jia, Junteng and Alwala, Kalyan Vasuden and Upasani, Kartikeya and Plawiak, Kate and Li, Ke and Heafield, Kenneth and Stone, Kevin and El-Arini, Khalid and Iyer, Krithika and Malik, Kshitiz and Chiu, Kuenley and Bhalla, Kunal and Rantala-Yeary, Lauren and van der Maaten, Laurens and Chen, Lawrence and Tan, Liang and Jenkins, Liz and Martin, Louis and Madaan, Lovish and Malo, Lubo and Blecher, Lukas and Landzaat, Lukas and de Oliveira, Luke and Muzzi, Madeline and Pasupuleti, Mahesh and Singh, Mannat and Paluri, Manohar and Kardas, Marcin and Oldham, Mathew and Rita, Mathieu and Pavlova, Maya and Kambadur, Melanie and Lewis, Mike and Si, Min and Singh, Mitesh Kumar and Hassan, Mona and Goyal, Naman and Torabi, Narjes and Bashlykov, Nikolay and Bogoychev, Nikolay and Chatterji, Niladri and Duchenne, Olivier and Çelebi, Onur and Alrassy, Patrick and Zhang, Pengchuan and Li, Pengwei and Vasic, Petar and Weng, Peter and Bhargava, Prajjwal and Dubal, Pratik and Krishnan, Praveen and Koura, Punit Singh and Xu, Puxin and He, Qing and Dong, Qingxiao and Srinivasan, Ragavan and Ganapathy, Raj and Calderer, Ramon and Cabral, Ricardo Silveira and Stojnic, Robert and Raileanu, Roberta and Girdhar, Rohit and Patel, Rohit and Sauvestre, Romain and Polidoro, Ronnie and Sumbaly, Roshan and Taylor, Ross and Silva, Ruan and Hou, Rui and Wang, Rui and Hosseini, Saghar and Chennabasappa, Sahana and Singh, Sanjay and Bell, Sean and Kim, Seohyun Sonia and Edunov, Sergey and Nie, Shaoliang and Narang, Sharan and Raparthy, Sharath and Shen, Sheng and Wan, Shengye and Bhosale, Shruti and Zhang, Shun and Vandenhende, Simon and Batra, Soumya and Whitman, Spencer and Sootla, Sten and Collot, Stephane and Gururangan, Suchin and Borodinsky, Sydney and Herman, Tamar and Fowler, Tara and Sheasha, Tarek and Georgiou, Thomas and Scialom, Thomas and Speckbacher, Tobias and Mihaylov, Todor and Xiao, Tong and Karn, Ujjwal and Goswami, Vedanuj and Gupta, Vibhor and Ramanathan, Vignesh and Kerkez, Viktor and Gonguet, Vincent and Do, Virginie and Vogeti, Vish and Petrovic, Vladan and Chu, Weiwei and Xiong, Wenhan and Fu, Wenyin and Meers, Whitney and Martinet, Xavier and Wang, Xiaodong and Tan, Xiaoqing Ellen and Xie, Xinfeng and Jia, Xuchao and Wang, Xuewei and Goldschlag, Yaelle and Gaur, Yashesh and Babaei, Yasmine and Wen, Yi and Song, Yiwen and Zhang, Yuchen and Li, Yue and Mao, Yuning and Coudert, Zacharie Delpierre and Yan, Zheng and Chen, Zhengxing and Papakipos, Zoe and Singh, Aaditya and Grattafiori, Aaron and Jain, Abha and Kelsey, Adam and Shajnfeld, Adam and Gangidi, Adithya and Victoria, Adolfo and Goldstand, Ahuva and Menon, Ajay and Sharma, Ajay and Boesenberg, Alex and Vaughan, Alex and Baevski, Alexei and Feinstein, Allie and Kallet, Amanda and Sangani, Amit and Yunus, Anam and Lupu, Andrei and Alvarado, Andres and Caples, Andrew and Gu, Andrew and Ho, Andrew and Poulton, Andrew and Ryan, Andrew and Ramchandani, Ankit and Franco, Annie and Saraf, Aparajita and Chowdhury, Arkabandhu and Gabriel, Ashley and Bharambe, Ashwin and Eisenman, Assaf and Yazdan, Azadeh and James, Beau and Maurer, Ben and Leonhardi, Benjamin and Huang, Bernie and Loyd, Beth and Paola, Beto De and Paranjape, Bhargavi and Liu, Bing and Wu, Bo and Ni, Boyu and Hancock, Braden and Wasti, Bram and Spence, Brandon and Stojkovic, Brani and Gamido, Brian and Montalvo, Britt and Parker, Carl and Burton, Carly and Mejia, Catalina and Wang, Changhan and Kim, Changkyu and Zhou, Chao and Hu, Chester and Chu, Ching-Hsiang and Cai, Chris and Tindal, Chris and Feichtenhofer, Christoph and Civin, Damon and Beaty, Dana and Kreymer, Daniel and Li, Daniel and Wyatt, Danny and Adkins, David and Xu, David and Testuggine, Davide and David, Delia and Parikh, Devi and Liskovich, Diana and Foss, Didem and Wang, Dingkang and Le, Duc and Holland, Dustin and Dowling, Edward and Jamil, Eissa and Montgomery, Elaine and Presani, Eleonora and Hahn, Emily and Wood, Emily and Brinkman, Erik and Arcaute, Esteban and Dunbar, Evan and Smothers, Evan and Sun, Fei and Kreuk, Felix and Tian, Feng and Ozgenel, Firat and Caggioni, Francesco and Guzmán, Francisco and Kanayet, Frank and Seide, Frank and Florez, Gabriela Medina and Schwarz, Gabriella and Badeer, Gada and Swee, Georgia and Halpern, Gil and Thattai, Govind and Herman, Grant and Sizov, Grigory and Guangyi and Zhang and Lakshminarayanan, Guna and Shojanazeri, Hamid and Zou, Han and Wang, Hannah and Zha, Hanwen and Habeeb, Haroun and Rudolph, Harrison and Suk, Helen and Aspegren, Henry and Goldman, Hunter and Damlaj, Ibrahim and Molybog, Igor and Tufanov, Igor and Veliche, Irina-Elena and Gat, Itai and Weissman, Jake and Geboski, James and Kohli, James and Asher, Japhet and Gaya, Jean-Baptiste and Marcus, Jeff and Tang, Jeff and Chan, Jennifer and Zhen, Jenny and Reizenstein, Jeremy and Teboul, Jeremy and Zhong, Jessica and Jin, Jian and Yang, Jingyi and Cummings, Joe and Carvill, Jon and Shepard, Jon and McPhie, Jonathan and Torres, Jonathan and Ginsburg, Josh and Wang, Junjie and Wu, Kai and U, Kam Hou and Saxena, Karan and Prasad, Karthik and Khandelwal, Kartikay and Zand, Katayoun and Matosich, Kathy and Veeraraghavan, Kaushik and Michelena, Kelly and Li, Keqian and Huang, Kun and Chawla, Kunal and Lakhotia, Kushal and Huang, Kyle and Chen, Lailin and Garg, Lakshya and A, Lavender and Silva, Leandro and Bell, Lee and Zhang, Lei and Guo, Liangpeng and Yu, Licheng and Moshkovich, Liron and Wehrstedt, Luca and Khabsa, Madian and Avalani, Manav and Bhatt, Manish and Tsimpoukelli, Maria and Mankus, Martynas and Hasson, Matan and Lennie, Matthew and Reso, Matthias and Groshev, Maxim and Naumov, Maxim and Lathi, Maya and Keneally, Meghan and Seltzer, Michael L. and Valko, Michal and Restrepo, Michelle and Patel, Mihir and Vyatskov, Mik and Samvelyan, Mikayel and Clark, Mike and Macey, Mike and Wang, Mike and Hermoso, Miquel Jubert and Metanat, Mo and Rastegari, Mohammad and Bansal, Munish and Santhanam, Nandhini and Parks, Natascha and White, Natasha and Bawa, Navyata and Singhal, Nayan and Egebo, Nick and Usunier, Nicolas and Laptev, Nikolay Pavlovich and Dong, Ning and Zhang, Ning and Cheng, Norman and Chernoguz, Oleg and Hart, Olivia and Salpekar, Omkar and Kalinli, Ozlem and Kent, Parkin and Parekh, Parth and Saab, Paul and Balaji, Pavan and Rittner, Pedro and Bontrager, Philip and Roux, Pierre and Dollar, Piotr and Zvyagina, Polina and Ratanchandani, Prashant and Yuvraj, Pritish and Liang, Qian and Alao, Rachad and Rodriguez, Rachel and Ayub, Rafi and Murthy, Raghotham and Nayani, Raghu and Mitra, Rahul and Li, Raymond and Hogan, Rebekkah and Battey, Robin and Wang, Rocky and Maheswari, Rohan and Howes, Russ and Rinott, Ruty and Bondu, Sai Jayesh and Datta, Samyak and Chugh, Sara and Hunt, Sara and Dhillon, Sargun and Sidorov, Sasha and Pan, Satadru and Verma, Saurabh and Yamamoto, Seiji and Ramaswamy, Sharadh and Lindsay, Shaun and Lindsay, Shaun and Feng, Sheng and Lin, Shenghao and Zha, Shengxin Cindy and Shankar, Shiva and Zhang, Shuqiang and Zhang, Shuqiang and Wang, Sinong and Agarwal, Sneha and Sajuyigbe, Soji and Chintala, Soumith and Max, Stephanie and Chen, Stephen and Kehoe, Steve and Satterfield, Steve and Govindaprasad, Sudarshan and Gupta, Sumit and Cho, Sungmin and Virk, Sunny and Subramanian, Suraj and Choudhury, Sy and Goldman, Sydney and Remez, Tal and Glaser, Tamar and Best, Tamara and Kohler, Thilo and Robinson, Thomas and Li, Tianhe and Zhang, Tianjun and Matthews, Tim and Chou, Timothy and Shaked, Tzook and Vontimitta, Varun and Ajayi, Victoria and Montanez, Victoria and Mohan, Vijai and Kumar, Vinay Satish and Mangla, Vishal and Albiero, Vítor and Ionescu, Vlad and Poenaru, Vlad and Mihailescu, Vlad Tiberiu and Ivanov, Vladimir and Li, Wei and Wang, Wenchen and Jiang, Wenwen and Bouaziz, Wes and Constable, Will and Tang, Xiaocheng and Wang, Xiaofang and Wu, Xiaojian and Wang, Xiaolan and Xia, Xide and Wu, Xilun and Gao, Xinbo and Chen, Yanjun and Hu, Ye and Jia, Ye and Qi, Ye and Li, Yenda and Zhang, Yilin and Zhang, Ying and Adi, Yossi and Nam, Youngjin and Yu and Wang and Hao, Yuchen and Qian, Yundi and He, Yuzi and Rait, Zach and DeVito, Zachary and Rosnbrick, Zef and Wen, Zhaoduo and Yang, Zhenyu and Zhao, Zhiwei},
  description = {The {Llama} 3 Herd of Models},
  journal = {arXiv preprint arXiv:2407.21783},
  title = {The Llama 3 Herd of Models},
  year = 2024
}

@article{penedo2023refinedweb,
  title   = {The {RefinedWeb} Dataset for {Falcon LLM}: Outperforming Curated Corpora with Web Data, and Web Data Only},
  author  = {Penedo, Guilherme and Malartic, Quentin and Hesslow, Daniel and Cojocaru, Ruxandra and Cappelli, Alessandro and Alobeidli, Hamza and Pannier, Baptiste and Almazrouei, Ebtesam and Launay, Julien},
  journal = {arXiv preprint arXiv:2306.01116},
  year    = {2023}
}

@inproceedings{wenzek2020ccnet,
  title     = {{CCNet}: Extracting High Quality Monolingual Datasets from Web Crawl Data},
  author    = {Wenzek, Guillaume and Lachaux, Marie-Anne and Conneau, Alexis and Chaudhary, Vishrav and Guzmán, Francisco and Joulin, Armand and Grave, Edouard},
  booktitle = {Proceedings of the Twelfth Language Resources and Evaluation Conference},
  pages     = {4003--4012},
  year      = {2020}
}

@article{tirumala2023d4,
  title   = {{D4}: Improving {LLM} Pretraining via Document De-Duplication and Diversification},
  author  = {Tirumala, Kushal and Simig, Daniel and Aghajanyan, Armen and Morcos, Ari S.},
  journal = {arXiv preprint arXiv:2308.12284},
  year    = {2023}
}

@misc{soboleva2023slimpajama,
  title        = {{SlimPajama}: A 627B Token Cleaned and Deduplicated Version of {RedPajama}},
  author       = {Soboleva, Daria and Al-Khateeb, Faisal and Myers, Robert and Steeves, Jacob R. and Hestness, Joel and Dey, Nolan},
  year         = {2023},
  howpublished = {Cerebras Blog}
}

@inproceedings{hendrycks2021mmlu,
  title={Measuring Massive Multitask Language Understanding},
  author={Hendrycks, Dan and Burns, Collin and Basart, Steven and Zou, Andy and Mazeika, Mantas and Song, Dawn and Steinhardt, Jacob},
  booktitle={Proceedings of the International Conference on Learning Representations (ICLR)},
  year={2021}
}

@article{clark2018arc,
  title={Think you have Solved Question Answering? {Try ARC}, the {AI2} Reasoning Challenge},
  author={Clark, Peter and Cowhey, Isaac and Etzioni, Oren and Khot, Tushar and Sabharwal, Ashish and Schoenick, Carissa and Tafjord, Oyvind},
  journal={arXiv preprint arXiv:1803.05457},
  year={2018}
}

@inproceedings{bandarkar2024belebele,
  title={The {Belebele} Benchmark: a Parallel Reading Comprehension Dataset in 122 Language Variants},
  author={Bandarkar, Lucas and Liang, Davis and Muller, Benjamin and Artetxe, Mikel and Shukla, Satya Narayan and Husa, Donald and Goyal, Naman and Krishnan, Abhinandan and Zettlemoyer, Luke and Khabsa, Madian},
  booktitle={Proceedings of the 62nd Annual Meeting of the Association for Computational Linguistics (ACL)},
  pages={749--775},
  year={2024}
}

@inproceedings{conneau2018xnli,
  title={{XNLI}: Evaluating Cross-lingual Sentence Representations},
  author={Conneau, Alexis and Rinott, Ruty and Lample, Guillaume and Williams, Adina and Bowman, Samuel R. and Schwenk, Holger and Stoyanov, Veselin},
  booktitle={Proceedings of the 2018 Conference on Empirical Methods in Natural Language Processing (EMNLP)},
  pages={2475--2485},
  year={2018}
}

@inproceedings{zellers2019hellaswag,
  title={{HellaSwag}: Can a Machine Really Finish Your Sentence?},
  author={Zellers, Rowan and Holtzman, Ari and Bisk, Yonatan and Farhadi, Ali and Choi, Yejin},
  booktitle={Proceedings of the 57th Annual Meeting of the Association for Computational Linguistics (ACL)},
  pages={4791--4800},
  year={2019}
}

@inproceedings{bisk2020piqa,
  title={{PIQA}: Reasoning about Physical Commonsense in Natural Language},
  author={Bisk, Yonatan and Zellers, Rowan and Le Bras, Ronan and Gao, Jianfeng and Choi, Yejin},
  booktitle={Proceedings of the AAAI Conference on Artificial Intelligence},
  volume={34},
  number={05},
  pages={7432--7439},
  year={2020}
}

@inproceedings{ponti2020xcopa,
  title={{XCOPA}: A Multilingual Dataset for Causal Commonsense Reasoning},
  author={Ponti, Edoardo Maria and Glava{\v{s}}, Goran and Majewska, Olga and Liu, Qianchu and Vuli{\'c}, Ivan and Korhonen, Anna},
  booktitle={Proceedings of the 2020 Conference on Empirical Methods in Natural Language Processing (EMNLP)},
  pages={2362--2376},
  year={2020}
}

@inproceedings{lewis2020mlqa,
  title={{MLQA}: Evaluating Cross-lingual Extractive Question Answering},
  author={Lewis, Patrick and Oguz, Barlas and Rinott, Ruty and Riedel, Sebastian and Schwenk, Holger},
  booktitle={Proceedings of the 58th Annual Meeting of the Association for Computational Linguistics (ACL)},
  pages={7315--7330},
  year={2020}
}

@article{clark2020tydiqa,
  title={{TyDi QA}: A Benchmark for Information-Seeking Question Answering in Typologically Diverse Languages},
  author={Clark, Jonathan H. and Choi, Eunsol and Collins, Michael and Garrette, Dan and Kwiatkowski, Tom and Nikolaev, Vitaly and Palomaki, Jennimaria},
  journal={Transactions of the Association for Computational Linguistics},
  volume={8},
  pages={454--470},
  year={2020}
}

@inproceedings{lin2022fewshot,
  title={Few-shot Learning with Multilingual Generative Language Models},
  author={Lin, Xi Victoria and Mihaylov, Todor and Artetxe, Mikel and Wang, Tianlu and Chen, Shuohui and Simig, Daniel and Ott, Myle and Goyal, Naman and Bhosale, Shruti and Du, Jingfei and Pasunuru, Ramakanth and Shleifer, Sam and Koura, Punit Singh and Chaudhary, Vishrav and O'Horo, Brian and Wang, Jeff and Zettlemoyer, Luke and Kozareva, Zornitsa and Diab, Mona and Stoyanov, Veselin and Li, Xian},
  booktitle={Proceedings of the 2022 Conference on Empirical Methods in Natural Language Processing (EMNLP)},
  pages={9019--9052},
  year={2022}
}

@inproceedings{artetxe2020xquad,
  title={On the Cross-lingual Transferability of Monolingual Representations},
  author={Artetxe, Mikel and Ruder, Sebastian and Yogatama, Dani},
  booktitle={Proceedings of the 58th Annual Meeting of the Association for Computational Linguistics (ACL)},
  pages={4623--4637},
  year={2020}
}

@inproceedings{hardalov2020exams,
  title={{EXAMS}: A Multi-subject High School Examinations Dataset for Cross-lingual and Multilingual Question Answering},
  author={Hardalov, Momchil and Mihaylov, Todor and Dimitrov, Dimitar and Zlatkova, Yordanka and Momchev, Preslav and Simov, Ivan and Nakov, Preslav},
  booktitle={Proceedings of the 2020 Conference on Empirical Methods in Natural Language Processing (EMNLP)},
  pages={5427--5444},
  year={2020}
}

@inproceedings{lai2017race,
  title={{RACE}: Large-scale Reading Comprehension Dataset From Examinations},
  author={Lai, Guokun and Xie, Qizhe and Liu, Hanxiao and Yang, Yiming and Hovy, Eduard},
  booktitle={Proceedings of the 2017 Conference on Empirical Methods in Natural Language Processing (EMNLP)},
  pages={785--794},
  year={2017}
}

@inproceedings{welbl2017sciq,
  title={Crowdsourcing Multiple Choice Science Questions},
  author={Welbl, Johannes and Liu, Nelson F. and Gardner, Matt},
  booktitle={Proceedings of the 3rd Workshop on Noisy User-generated Text},
  pages={94--106},
  year={2017}
}

@inproceedings{lin2021xcodah,
  title={Common Sense Beyond {English}: Evaluating and Improving Multilingual Language Models for Commonsense Reasoning},
  author={Lin, Bill Yuchen and Lee, Seyeon and Qiao, Xiaoyang and Ren, Xiang},
  booktitle={Proceedings of the 59th Annual Meeting of the Association for Computational Linguistics (ACL)},
  pages={1274--1287},
  year={2021}
}

@inproceedings{doddapaneni2023indicqa,
  title={Towards Leaving No {Indic} Language Behind: Building Monolingual Corpora, Benchmark and Models for {Indic} Languages},
  author={Doddapaneni, Sumanth and Aralikatte, Rahul and Ramesh, Gowtham and Goyal, Shreya and Khapra, Mitesh M. and Kunchukuttan, Anoop and Kumar, Pratyush},
  booktitle={Proceedings of the 61st Annual Meeting of the Association for Computational Linguistics (ACL)},
  year={2023}
}

@article{gemmateam2024gemma2improvingopen,
      title={Gemma 2: Improving Open Language Models at a Practical Size}, 
      author={Gemma Team and Morgane Riviere and Shreya Pathak and Pier Giuseppe Sessa and Cassidy Hardin and Surya Bhupatiraju and Léonard Hussenot and Thomas Mesnard and Bobak Shahriari and Alexandre Ramé and Johan Ferret and Peter Liu and Pouya Tafti and Abe Friesen and Michelle Casbon and Sabela Ramos and Ravin Kumar and Charline Le Lan and Sammy Jerome and Anton Tsitsulin and Nino Vieillard and Piotr Stanczyk and Sertan Girgin and Nikola Momchev and Matt Hoffman and Shantanu Thakoor and Jean-Bastien Grill and Behnam Neyshabur and Olivier Bachem and Alanna Walton and Aliaksei Severyn and Alicia Parrish and Aliya Ahmad and Allen Hutchison and Alvin Abdagic and Amanda Carl and Amy Shen and Andy Brock and Andy Coenen and Anthony Laforge and Antonia Paterson and Ben Bastian and Bilal Piot and Bo Wu and Brandon Royal and Charlie Chen and Chintu Kumar and Chris Perry and Chris Welty and Christopher A. Choquette-Choo and Danila Sinopalnikov and David Weinberger and Dimple Vijaykumar and Dominika Rogozińska and Dustin Herbison and Elisa Bandy and Emma Wang and Eric Noland and Erica Moreira and Evan Senter and Evgenii Eltyshev and Francesco Visin and Gabriel Rasskin and Gary Wei and Glenn Cameron and Gus Martins and Hadi Hashemi and Hanna Klimczak-Plucińska and Harleen Batra and Harsh Dhand and Ivan Nardini and Jacinda Mein and Jack Zhou and James Svensson and Jeff Stanway and Jetha Chan and Jin Peng Zhou and Joana Carrasqueira and Joana Iljazi and Jocelyn Becker and Joe Fernandez and Joost van Amersfoort and Josh Gordon and Josh Lipschultz and Josh Newlan and Ju-yeong Ji and Kareem Mohamed and Kartikeya Badola and Kat Black and Katie Millican and Keelin McDonell and Kelvin Nguyen and Kiranbir Sodhia and Kish Greene and Lars Lowe Sjoesund and Lauren Usui and Laurent Sifre and Lena Heuermann and Leticia Lago and Lilly McNealus and Livio Baldini Soares and Logan Kilpatrick and Lucas Dixon and Luciano Martins and Machel Reid and Manvinder Singh and Mark Iverson and Martin Görner and Mat Velloso and Mateo Wirth and Matt Davidow and Matt Miller and Matthew Rahtz and Matthew Watson and Meg Risdal and Mehran Kazemi and Michael Moynihan and Ming Zhang and Minsuk Kahng and Minwoo Park and Mofi Rahman and Mohit Khatwani and Natalie Dao and Nenshad Bardoliwalla and Nesh Devanathan and Neta Dumai and Nilay Chauhan and Oscar Wahltinez and Pankil Botarda and Parker Barnes and Paul Barham and Paul Michel and Pengchong Jin and Petko Georgiev and Phil Culliton and Pradeep Kuppala and Ramona Comanescu and Ramona Merhej and Reena Jana and Reza Ardeshir Rokni and Rishabh Agarwal and Ryan Mullins and Samaneh Saadat and Sara Mc Carthy and Sarah Cogan and Sarah Perrin and Sébastien M. R. Arnold and Sebastian Krause and Shengyang Dai and Shruti Garg and Shruti Sheth and Sue Ronstrom and Susan Chan and Timothy Jordan and Ting Yu and Tom Eccles and Tom Hennigan and Tomas Kocisky and Tulsee Doshi and Vihan Jain and Vikas Yadav and Vilobh Meshram and Vishal Dharmadhikari and Warren Barkley and Wei Wei and Wenming Ye and Woohyun Han and Woosuk Kwon and Xiang Xu and Zhe Shen and Zhitao Gong and Zichuan Wei and Victor Cotruta and Phoebe Kirk and Anand Rao and Minh Giang and Ludovic Peran and Tris Warkentin and Eli Collins and Joelle Barral and Zoubin Ghahramani and Raia Hadsell and D. Sculley and Jeanine Banks and Anca Dragan and Slav Petrov and Oriol Vinyals and Jeff Dean and Demis Hassabis and Koray Kavukcuoglu and Clement Farabet and Elena Buchatskaya and Sebastian Borgeaud and Noah Fiedel and Armand Joulin and Kathleen Kenealy and Robert Dadashi and Alek Andreev},
      year={2024},
      journal={arXiv preprint arXiv:2408.00118},
}

@inproceedings{lee2022deduplicating,
  title={Deduplicating Training Data Makes Language Models Better},
  author={Lee, Katherine and Ippolito, Daphne and Nystrom, Andrew and Zhang, Chiyuan and Eck, Douglas and Callison-Burch, Chris and Carlini, Nicholas},
  booktitle={Proceedings of the 60th Annual Meeting of the Association for Computational Linguistics (Volume 1: Long Papers)},
  pages={8424--8445},
  year={2022},
  address={Dublin, Ireland},
  publisher={Association for Computational Linguistics}
}

@article{messmer2025enhancingmultilingualllmpretraining,
      title={Enhancing Multilingual {LLM} Pretraining with Model-Based Data Selection}, 
      author={Bettina Messmer and Vinko Sabolčec and Martin Jaggi},
      year={2025},
      journal={arXiv preprint arXiv:2502.10361},
}

@article{fleiss1971measuring,
  title={Measuring Nominal Scale Agreement Among Many Raters},
  author={Fleiss, Joseph L.},
  journal={Psychological Bulletin},
  volume={76},
  number={5},
  pages={378--382},
  year={1971},
  publisher={American Psychological Association}
}

@article{krippendorff2011computing,
  title={Computing {K}rippendorff's Alpha-Reliability},
  author={Krippendorff, Klaus},
  journal={Departmental Papers (ASC)},
  pages={43},
  year={2011},
  publisher={University of Pennsylvania}
}

@article{dawid1979maximum,
  title={Maximum Likelihood Estimation of Observer Error-Rates Using the {EM} Algorithm},
  author={Dawid, A. Philip and Skene, Allan M.},
  journal={Journal of the Royal Statistical Society. Series C (Applied Statistics)},
  volume={28},
  number={1},
  pages={20--28},
  year={1979},
  publisher={Wiley}
}

@article{breiman1996bagging,
  title={Bagging Predictors},
  author={Breiman, Leo},
  journal={Machine Learning},
  volume={24},
  number={2},
  pages={123--140},
  year={1996},
  publisher={Springer}
}

@article{freund1997decision,
  title={A Decision-Theoretic Generalization of On-Line Learning and an Application to Boosting},
  author={Freund, Yoav and Schapire, Robert E.},
  journal={Journal of Computer and System Sciences},
  volume={55},
  number={1},
  pages={119--139},
  year={1997},
  publisher={Elsevier}
}

@article{hoffmann2022chinchilla,
      title={Training Compute-Optimal Large Language Models}, 
      author={Jordan Hoffmann and Sebastian Borgeaud and Arthur Mensch and Elena Buchatskaya and Trevor Cai and Eliza Rutherford and Diego de Las Casas and Lisa Anne Hendricks and Johannes Welbl and Aidan Clark and Tom Hennigan and Eric Noland and Katie Millican and George van den Driessche and Bogdan Damoc and Aurelia Guy and Simon Osindero and Karen Simonyan and Erich Elsen and Jack W. Rae and Oriol Vinyals and Laurent Sifre},
      year={2022},
      journal={arXiv preprint arXiv:2203.15556},
}

@inproceedings{ortiz2019oscar,
  title     = {Asynchronous Pipeline for Processing Huge Corpora on Medium to Low Resource Infrastructures},
  author    = {Ortiz Su{\'a}rez, Pedro Javier and Sagot, Beno{\^\i}t and Romary, Laurent},
  booktitle = {Proceedings of the Workshop on Challenges in the Management of Large Corpora (CMLC-7)},
  pages     = {9--16},
  year      = {2019}
}

@misc{together2023redpajama,
  title        = {{RedPajama}: An Open Source Recipe to Reproduce {LLaMA} Training Dataset},
  author       = {{Together Computer}},
  year         = {2023},
  howpublished = {GitHub Repository},
  url          = {https://github.com/togethercomputer/RedPajama-Data}
}

@article{khan2024sangraha,
  title   = {Sangraha: A Large-scale, High-quality and Diverse Corpus for {Indic} Languages},
  author  = {Khan, Mohammed Safi Ur Rahman and Kunchukuttan, Anoop and Khapra, Mitesh M. and Kumar, Pratyush},
  journal = {arXiv preprint arXiv:2403.06350},
  year    = {2024}
}

@misc{vngrs2024web,
  title        = {{VNGRS-Web}: A {Turkish} Web Corpus},
  author       = {{VNGRS}},
  year         = {2024},
  howpublished = {Hugging Face Datasets},
  url          = {https://huggingface.co/datasets/vngrs-ai/vngrs-web-corpus}
}
\newpage
\appendix

\section{Documents by Sources}
\label{sec:sources_appendix}

\begin{table*}[t]
\centering
\caption{Complete source-level statistics for all Matched and MinHash-deduplicated corpora.}
\label{tab:appendix_full_sources}
\small
\begin{tabular}{@{}l l r r r r@{}}
\toprule
 & & \multicolumn{2}{c}{\textbf{Matched}} & \multicolumn{2}{c}{\textbf{MinHash}} \\
\cmidrule(lr){3-4} \cmidrule(lr){5-6}
\textbf{Language} & \textbf{Source} & \textbf{Tokens} & \textbf{Documents} & \textbf{Tokens} & \textbf{Documents} \\
\midrule
Arabic
 & CulturaX & 41,873,154,858 & 34,406,699 & 42,101,381,105 & 44,237,981 \\
 & ArabicWeb24 & 10,341,870,035 & 10,298,259 & 35,387,940,667 & 34,538,428 \\
 & HPLT 2.0 & 26,299,111,184 & 23,509,861 & 34,662,907,407 & 34,005,137 \\
 & FineWeb-2 & 36,443,315,158 & 31,995,522 & 27,468,352,797 & 32,048,536 \\
 & C4 & 27,628,339,664 & 23,860,944 & 22,483,177,493 & 26,856,527 \\
 & ClusterLab & 615,093,350 & 321,531 & 9,477,371,984 & 6,527,100 \\
 & FinePDFs & 25,212,064 & 5,551 & 6,251,291,619 & 669,532 \\
 & \textbf{Total} & \textbf{54,077,826,212} & \textbf{47,913,611} & \textbf{177,832,423,072} & \textbf{178,883,241} \\
\midrule
Turkish
 & HPLT 2.0 & 31,405,049,730 & 37,011,815 & 46,031,725,288 & 53,713,070 \\
 & FineWeb-2 & 39,069,031,983 & 46,477,799 & 41,872,725,965 & 54,528,717 \\
 & CulturaX & 38,930,592,123 & 45,150,639 & 35,773,741,573 & 47,867,770 \\
 & C4 & 32,064,877,058 & 37,404,178 & 25,278,653,392 & 36,459,586 \\
 & VNGRS-Web & 21,245,667,673 & 24,190,883 & 18,675,771,237 & 26,503,327 \\
 & \textbf{Total} & \textbf{55,956,064,858} & \textbf{67,640,864} & \textbf{167,632,617,455} & \textbf{219,072,470} \\
\midrule
Hindi
 & FineWeb-2 & 15,693,519,629 & 12,303,399 & 20,034,066,122 & 17,149,268 \\
 & CulturaX & 21,538,268,621 & 14,747,150 & 16,649,950,565 & 11,512,850 \\
 & Sangraha (unverified) & 18,258,210,651 & 13,088,202 & 11,512,925,283 & 8,940,999 \\
 & HPLT 2.0 & 10,235,635,825 & 6,724,779 & 10,226,668,227 & 6,674,725 \\
 & Sangraha (verified) & 2,976,801,145 & 2,722,016 & 10,060,767,856 & 9,050,884 \\
 & C4 & 10,661,073,895 & 7,148,057 & 7,703,559,509 & 6,250,938 \\
 & \textbf{Total} & \textbf{27,145,973,874} & \textbf{19,767,791} & \textbf{76,187,937,562} & \textbf{59,579,664} \\
\bottomrule
\end{tabular}
\end{table*}

Table~\ref{tab:appendix_full_sources} provides a complete breakdown of token and document counts by source for both the matched and MinHash-deduplicated corpora. These statistics reveal how each source contributes to the final datasets and highlight the asymmetry between raw volume and cross-source validation.

For Arabic, CulturaX contributes the most tokens to the matched subset (41.9B), despite ArabicWeb24 providing the largest share of unique content in the MinHash pool. This pattern reflects our methodology: sources with high pairwise overlap contribute disproportionately to the matched subset, while sources with unique content (such as FinePDFs and ArabicWeb24) dominate the MinHash pool. The matched subset thus emphasizes content that multiple pipelines independently retained, while the MinHash pool preserves the full diversity of available sources.

Similar patterns emerge for Turkish and Hindi. In Turkish, the three largest sources (HPLT~2.0, FineWeb-2, and CulturaX) contribute roughly equal shares to the matched subset, reflecting substantial three-way overlap among Common Crawl derivatives. In Hindi, CulturaX and the Sangraha corpora show high mutual overlap, with CulturaX contributing the largest matched share despite FineWeb-2 providing more unique content overall.

These distributions inform practical decisions about corpus construction. Practitioners seeking maximum diversity should use the full MinHash pool; those prioritizing cross-validated quality should use the matched subset. The per-document source counts we release enable intermediate configurations.

\section{Detailed Pairwise Overlaps}
\label{sec:pairwise_appendix}

We report complete pairwise overlap statistics for all source combinations. These tables extend Figure~\ref{fig:pairwise} from the main text, providing exact token and document counts for each source pair.

\begin{table}[t]
\centering
\caption{Largest pairwise source overlaps in matched corpora, ranked by token count.}
\label{tab:top_pairwise}
\small
\begin{tabular}{@{}l l l r r@{}}
\toprule
\textbf{Language} & \textbf{Source A} & \textbf{Source B} & \textbf{Tokens} & \textbf{Docs} \\
\midrule
Turkish & FineWeb-2 & HPLT 2.0 & 9.19B & 13.3M \\
Arabic  & C4 & CulturaX & 9.16B & 8.9M \\
Arabic  & CulturaX & FineWeb-2 & 6.11B & 4.6M \\
Arabic  & FineWeb-2 & HPLT 2.0 & 5.62B & 6.6M \\
Turkish & C4 & CulturaX & 5.23B & 7.9M \\
Turkish & CulturaX & FineWeb-2 & 5.18B & 6.1M \\
Hindi   & CulturaX & Sangraha & 4.05B & 2.9M \\
\bottomrule
\end{tabular}
\end{table}

\begin{table}[t]
\centering
\caption{Pairwise source overlaps in Arabic pretraining corpora.}
\label{tab:appendix_arabic_pairwise}
\small
\begin{tabular}{@{}l l r r@{}}
\toprule
\textbf{Source A} & \textbf{Source B} & \textbf{Tokens} & \textbf{Docs} \\
\midrule
C4 & CulturaX & 9.16B & 8.93M \\
CulturaX & FineWeb-2 & 6.11B & 4.58M \\
FineWeb-2 & HPLT 2.0 & 5.62B & 6.59M \\
CulturaX & HPLT 2.0 & 2.76B & 1.89M \\
ArabicWeb24 & FineWeb-2 & 2.53B & 2.90M \\
ArabicWeb24 & HPLT 2.0 & 0.80B & 0.92M \\
C4 & FineWeb-2 & 0.66B & 0.58M \\
C4 & HPLT 2.0 & 0.55B & 0.48M \\
ArabicWeb24 & CulturaX & 0.41B & 0.44M \\
ClusterLab & CulturaX & 0.10B & 0.05M \\
\bottomrule
\end{tabular}
\end{table}

Table~\ref{tab:top_pairwise} ranks the largest pairwise overlaps across all languages. The dominance of Common Crawl derivatives is evident: the top overlaps involve FineWeb-2, HPLT~2.0, CulturaX, and C4, all of which process Common Crawl snapshots with different filtering pipelines. The largest single overlap (FineWeb-2 and HPLT~2.0 for Turkish, 9.19B tokens) represents content that two independent teams deemed worth retaining from the same underlying web snapshots.

Tables~\ref{tab:appendix_arabic_pairwise},~\ref{tab:appendix_turkish_pairwise}, and~\ref{tab:appendix_hindi_pairwise} provide language-specific breakdowns. Several patterns merit attention. First, overlap magnitude varies substantially by source pair: C4 and CulturaX share 9.16B Arabic tokens, while ArabicWeb24 and CulturaX share only 0.41B. Second, specialized sources show minimal overlap with general crawls: FinePDFs, which processes academic PDFs rather than web pages, shares negligible content with other Arabic sources. Third, overlap is not symmetric in relative terms: a source pair may represent a large fraction of one source but a small fraction of another, depending on their respective sizes.

These statistics motivate our cross-source matching approach. In fact, the substantial overlaps indicate that independent pipelines frequently retain the same content, providing the redundancy our method exploits. Moreover, the variation in overlap magnitude suggests that requiring two or more sources strikes a reasonable balance: stricter thresholds would exclude too much content, while looser thresholds would admit content validated by only marginal agreement.

\begin{table}[h]
\centering
\caption{Pairwise source overlaps in Turkish pretraining corpora.}
\label{tab:appendix_turkish_pairwise}
\small
\begin{tabular}{@{}l l r r@{}}
\toprule
\textbf{Source A} & \textbf{Source B} & \textbf{Tokens} & \textbf{Docs} \\
\midrule
FineWeb-2 & HPLT 2.0 & 9.19B & 13.31M \\
C4 & CulturaX & 5.23B & 7.90M \\
CulturaX & FineWeb-2 & 5.18B & 6.08M \\
C4 & VNGRS & 2.28B & 2.93M \\
CulturaX & HPLT 2.0 & 1.62B & 1.78M \\
C4 & HPLT 2.0 & 0.93B & 0.77M \\
C4 & FineWeb-2 & 0.76B & 0.80M \\
CulturaX & VNGRS & 0.36B & 0.53M \\
HPLT 2.0 & VNGRS & 0.29B & 0.39M \\
FineWeb-2 & VNGRS & 0.21B & 0.43M \\
\bottomrule
\end{tabular}
\end{table}

\begin{table}[h]
\centering
\caption{Pairwise source overlaps in Hindi pretraining corpora.}
\label{tab:appendix_hindi_pairwise}
\small
\begin{tabular}{@{}l l r r@{}}
\toprule
\textbf{Source A} & \textbf{Source B} & \textbf{Tokens} & \textbf{Docs} \\
\midrule
CulturaX & Sangraha (unverif.) & 4.05B & 2.89M \\
FineWeb-2 & HPLT 2.0 & 2.08B & 1.76M \\
C4 & CulturaX & 1.49B & 1.04M \\
FineWeb-2 & Sangraha (verif.) & 0.98B & 1.10M \\
CulturaX & FineWeb-2 & 0.54B & 0.41M \\
CulturaX & HPLT 2.0 & 0.49B & 0.23M \\
C4 & FineWeb-2 & 0.46B & 0.36M \\
C4 & HPLT 2.0 & 0.31B & 0.15M \\
FineWeb-2 & Sangraha (unverif.) & 0.31B & 0.38M \\
HPLT 2.0 & Sangraha (verif.) & 0.22B & 0.19M \\
\bottomrule
\end{tabular}
\end{table}

\section{Evaluation Suite}
\label{sec:eval_suite_appendix}

We evaluate models using language-specific benchmark suites that span four capability categories: general knowledge (GK), reading comprehension (RC), reasoning (RES), and natural language understanding (NLU). Table~\ref{tab:benchmarks} lists all benchmarks by language and category.

Our benchmark selection follows the FineWeb-2 methodology, prioritizing tasks that exhibit three properties: monotonic improvement with training compute, relatively low noise across random seeds, and consistent model ordering. So, we exclude benchmarks where random performance is indistinguishable from trained models at our scale, or where evaluation variance obscures genuine quality differences.

For each language, we include both multilingual benchmarks (XNLI, Belebele, HellaSwag) and language-specific evaluations where available. Arabic includes ARCD and SoQAL for reading comprehension; Turkish includes TQuADv2; Hindi includes IndicQA and IndicXCOPA. This combination ensures that our aggregate scores reflect both cross-lingual transfer and language-specific competence.

Table~\ref{tab:baselines} reports the random baseline for each task format. Multiple-choice tasks use uniform random selection as the baseline (0.25 for 4-way, 0.333 for 3-way, etc.), while generative question-answering tasks use a baseline of zero. Our rescaled scores (Section~\ref{evaluation_section}) normalize against these baselines, ensuring that aggregate scores are comparable across tasks with different numbers of answer choices.

\begin{table*}[h]
\centering
\caption{Evaluation benchmarks organized by language and category. GK: General Knowledge; RC: Reading Comprehension; RES: Reasoning; NLU: Natural Language Understanding.}
\label{tab:benchmarks}
\small
\begin{tabular}{@{}p{0.06\textwidth}p{0.06\textwidth}p{0.82\textwidth}@{}}
\toprule
\textbf{Lang.} & \textbf{Cat.} & \textbf{Benchmarks} \\
\midrule
Arabic
 & GK  & EXAMS~\citep{hardalov2020exams}, MMLU~\citep{hendrycks2021mmlu}, ARC~\citep{clark2018arc} \\
 & RC  & Belebele~\citep{bandarkar2024belebele}, SoQAL, RACE~\citep{lai2017race}, SciQ~\citep{welbl2017sciq}, MLQA~\citep{lewis2020mlqa}, TyDiQA~\citep{clark2020tydiqa}, ARCD \\
 & RES & X-CODAH~\citep{lin2021xcodah}, PIQA~\citep{bisk2020piqa}, X-CSQA \\
 & NLU & XNLI~\citep{conneau2018xnli}, HellaSwag~\citep{zellers2019hellaswag}, XStoryCloze~\citep{lin2022fewshot} \\
\midrule
Turkish
 & GK  & EXAMS, MMLU, ARC \\
 & RC  & Belebele, XCOPA~\citep{ponti2020xcopa}, TQuADv2, XQuAD~\citep{artetxe2020xquad} \\
 & RES & X-CODAH, PIQA, X-CSQA \\
 & NLU & XNLI, HellaSwag, XStoryCloze \\
\midrule
Hindi
 & GK  & MMLU, ARC \\
 & RC  & Belebele, IndicQA~\citep{doddapaneni2023indicqa}, XQuAD, MLQA \\
 & RES & IndicXCOPA, PIQA, X-CSQA \\
 & NLU & IndicNXNLI, XNLI, HellaSwag, XStoryCloze \\
\bottomrule
\end{tabular}
\end{table*}

\begin{table}[htb]
\centering
\caption{Random baseline values by task format. The baseline corresponds to the expected accuracy under uniform random selection.}
\label{tab:baselines}
\small
\begin{tabular}{@{}cp{4.2cm}c@{}}
\toprule
\textbf{Choices} & \textbf{Tasks} & \textbf{Baseline} \\
\midrule
4 & MMLU, EXAMS, ARC, Belebele, SoQAL, RACE, SciQ, X-CODAH, HellaSwag, XCOPA, IndicXCOPA & 0.25 \\
3 & XNLI, IndicNXNLI & 0.333 \\
2 & PIQA, XStoryCloze & 0.50 \\
5 & X-CSQA & 0.20 \\
Gen. & MLQA, TyDiQA, ARCD, XQuAD, TQuADv2, IndicQA & 0.00 \\
\bottomrule
\end{tabular}
\end{table}

\section{Domain Distribution in AraMix-MinHash}
\label{sec:domain_appendix}

To characterize corpus composition beyond source provenance, we classify all documents in AraMix-MinHash using Nvidia's multilingual domain classifier (\texttt{nvidia/domain-classifier}). Table~\ref{tab:domain_minhash} reports the distribution across 26 categories for the MinHash-deduplicated subset.

The corpus exhibits broad coverage without excessive concentration in any single domain. People and Society contributes the largest share (14.8\%), followed by News (14.1\%) and Business and Industrial (10.7\%). Notably, 19 of 26 categories contain over 1 billion tokens each, indicating that the cross-source matching process does not systematically favor particular content types.

Several observations merit attention. First, Sensitive Subjects ranks fourth (8.5\%), reflecting the prevalence of religious and political content in Arabic web text. Second, Science remains underrepresented (1.1\%), a known limitation of web-crawled corpora that may warrant supplementation with academic sources for domain-specific applications. Third, the Adult category contains only 1.4B tokens (0.8\%), suggesting that quality filters across source datasets effectively remove such content before our aggregation step.

We include the predicted domain as metadata for each document in our released corpora, enabling practitioners to construct domain-weighted training mixtures or to exclude specific categories for their applications.

\begin{table}[ht]
\centering
\caption{Domain distribution of the MinHash-deduplicated Arabic corpus (178B tokens). Percentages indicate share of total tokens.}
\label{tab:domain_minhash}
\begin{tabular}{@{}lrr@{}}
\toprule
\textbf{Domain} & \textbf{Tokens (B)} & \textbf{\%} \\
\midrule
People and Society & 26.4 & 14.8 \\
News & 25.1 & 14.1 \\
Business and Industrial & 19.1 & 10.7 \\
Sensitive Subjects & 15.1 & 8.5 \\
Health & 9.6 & 5.4 \\
Finance & 8.6 & 4.8 \\
Sports & 8.1 & 4.6 \\
Arts and Entertainment & 6.7 & 3.8 \\
Books and Literature & 6.5 & 3.7 \\
Jobs and Education & 6.5 & 3.7 \\
Food and Drink & 5.7 & 3.2 \\
Law and Government & 5.5 & 3.1 \\
Home and Garden & 5.4 & 3.0 \\
Travel and Transportation & 5.2 & 2.9 \\
Beauty and Fitness & 4.2 & 2.4 \\
Computers and Electronics & 3.5 & 2.0 \\
Internet and Telecom & 3.3 & 1.9 \\
Science & 2.0 & 1.1 \\
Autos and Vehicles & 2.0 & 1.1 \\
Games & 2.0 & 1.1 \\
Hobbies and Leisure & 1.4 & 0.8 \\
Pets and Animals & 1.4 & 0.8 \\
Shopping & 1.4 & 0.8 \\
Adult & 1.4 & 0.8 \\
Real Estate & 1.1 & 0.6 \\
Online Communities & 0.7 & 0.4 \\
\bottomrule
\end{tabular}
\end{table}

\subsection{Representative Examples}

To illustrate the content captured by cross-source matching, we present representative documents from three domains. These examples were selected from the matched subset (appearing in two or more sources) and are translated from Arabic for accessibility.

\noindent\textbf{People and Society.}
A community news item shared on social media by residents of Ufa, the capital of Bashkortostan, reporting the installation of a new dome on the Al-Rahim mosque to replace one damaged by a windstorm. The text quotes the head of the Muslim Religious Administration of Bashkortostan announcing the installation time via Telegram. This example typifies the local news and community updates that comprise much of the People and Society category.

\noindent\textbf{Books and Literature.}
A book review of an Arabic translation of Ernest Hemingway's short stories, translated by Rana Daoud and Sanaa Muhammad. The review discusses Hemingway's psychological style, his minimal physical description of characters, and his focus on universal themes. It references specific stories including ``Hills Like White Elephants'' and ``The Snows of Kilimanjaro.'' Literary criticism and book reviews appear frequently in this category, often exhibiting the thoughtful prose that cross-source agreement tends to surface.

\noindent\textbf{Science.}
Coverage of a paleoanthropological discovery in Ethiopia: a 3.8-million-year-old skull of early human ancestors that challenges prevailing theories of human evolution. The article references publication in Nature and names the discovering researcher, Professor Yohannes Haile-Selassie. Science journalism of this quality is relatively rare in the corpus, but cross-source matching helps identify the higher-quality examples that multiple pipelines retained.

These examples demonstrate that cross-source agreement surfaces substantive content across diverse domains, from local community news to literary criticism to science journalism. The common thread is coherent, informative text that multiple independent filtering pipelines deemed worth preserving.

\section{Limitations}
\label{sef:limitations_appendix}

\textbf{Scale.} Our experiments train 1.46B parameter models on 29B tokens. While this is modest by contemporary standards, it follows the experimental protocol established by FineWeb-2 for corpus quality comparisons, enabling direct comparison with their reported baselines. Whether our findings transfer to larger model scales (7B+ parameters) or longer training runs remains an open question, though we note that data quality effects typically persist or amplify at scale~\citep{hoffmann2022chinchilla}.

\textbf{Languages.} We construct corpora for four languages, but present full training experiments for three (Arabic, Hindi, Turkish).

\textbf{Threshold selection.} We use a fixed threshold of 2+ sources for the matched subset. The optimal threshold likely varies by language and depends on the number and diversity of available sources. We release per-document source counts to enable practitioners to tune this threshold for their use case. Tuning this parameter could only lead to either improved or similar performance.

\textbf{Complementary methods.} We compare against single-source baselines and one model-based approach, but do not exhaustively benchmark against all data selection methods (DSIR, perplexity filtering, etc.). Cross-source matching is complementary to these approaches, exploring hybrid pipelines is promising future work.

\end{document}